%% file: AAAI_main.tex
\def\year{2021}\relax
\documentclass[letterpaper]{article} % DO NOT CHANGE THIS
\usepackage{aaai21}  % DO NOT CHANGE THIS
\usepackage{times}  % DO NOT CHANGE THIS
\usepackage{helvet} % DO NOT CHANGE THIS
\usepackage{courier}  % DO NOT CHANGE THIS
\usepackage[hyphens]{url}  % DO NOT CHANGE THIS
\usepackage{graphicx} % DO NOT CHANGE THIS
\usepackage{natbib}  % DO NOT CHANGE THIS AND DO NOT ADD ANY OPTIONS TO IT
\usepackage{caption} % DO NOT CHANGE THIS AND DO NOT ADD ANY OPTIONS TO IT
\usepackage{mathrsfs}
\usepackage{amsfonts}       % blackboard math symbols
\usepackage{multirow}
\usepackage{amsmath}
\usepackage{amssymb}
\usepackage{xcolor}
\usepackage{float}

\title{Addressing Class Imbalance in Federated Learning}

\author {
        Lixu Wang\textsuperscript{\rm 1},
        Shichao Xu\textsuperscript{\rm 1},
        Xiao Wang\textsuperscript{\rm 1},
        Qi Zhu\textsuperscript{\rm 1} \\
}
\affiliations {
    \textsuperscript{\rm 1} Northwestern University, Evanston, IL, USA \\
    \{lixuwang2025, shichaoxu2023\}@u.northwestern.edu, \{wangxiao, qzhu\}@northwestern.edu
}

\begin{document}
\maketitle

\begin{abstract}
Federated learning (FL) is a promising approach for training decentralized data located on local client devices while improving efficiency and privacy. However, the distribution and quantity of the training data on the clients' side may lead to significant challenges such as class imbalance and non-IID (non-independent and identically distributed) data, which could greatly impact the performance of the common model. While much effort has been devoted to helping FL models converge when encountering non-IID data, the imbalance issue has not been sufficiently addressed.
In particular, as FL training is executed by exchanging gradients in an encrypted form, the training data is not completely observable to either clients or server, and previous methods for class imbalance do not perform well for FL. 
Therefore, it is crucial to design new methods for detecting class imbalance in FL and mitigating its impact. 
In this work, we propose a monitoring scheme that can infer the composition of training data for each FL round, and design a new loss function -- \textbf{Ratio Loss} to mitigate the impact of the imbalance.
Our experiments demonstrate the importance of acknowledging class imbalance and taking measures as early as possible in FL training, and the effectiveness of our method in mitigating the impact. Our method is shown to significantly outperform previous methods, while maintaining client privacy.
\end{abstract}

\input{Introduction}
\input{Related_Work}
\input{Method}
\input{Evaluation}
\input{Conclusion}
\input{Acknowledgment}
\newpage
\input{Supplementary}
\newpage
\bibstyle{aaai}
\bibliography{reference}
\end{document}

%% file: Introduction.tex
\section{Introduction}
\label{intro}
The emergence of federated learning (FL) enables multiple devices to collaboratively learn a common model without the need to collect data directly from local devices. It reduces the resource consumption on the cloud and also enhances the client privacy. FL has seen promising applications in multiple domains, including mobile phones~\cite{hard2018federated, ramaswamy2019federated}, wearable devices~\cite{nguyen2018d}, autonomous vehicles~\cite{samarakoon2018distributed}, etc.

In standard FL, a random subset of clients will be selected in each iteration, who will upload their gradient updates to the central server. The server will then aggregate those updates and return the updated common model to all participants. 
During FL, one major challenge is that data owned by different clients comes from various sources and may contain their own preferences, and the resulting diversity may make the convergence of the global model challenging and slow. 
Moreover, the phenomenon of class imbalance happens frequently in practical scenarios, e.g., the number of patients diagnosed with different diseases varies greatly~\cite{rao2006data, dong2019semantic}, and people have different preferences when typing with G-board~\cite{ramaswamy2019federated} (a practical FL application proposed by Google).
When a model encounters class imbalance, samples of \emph{majority classes} account for a very large proportion of the overall training data, while those of \emph{minority classes} account for much less.
The direct impact of class imbalance is the reduction of classification accuracy on minority classes. In many practical cases, those minority classes play a much more important role beyond their proportion in data, e.g., wearable devices need to be more sensitive to abnormal heart rates than normal scenarios, and it is more important for G-board to predict SOS precisely than restaurant names. 

In the literature, a number of approaches have been proposed to address class imbalance, e.g., applying various data sampling techniques~\cite{jo2004class}, using generative augmentation to make up for the lack of minority samples~\cite{lee2016plankton, pouyanfar2018dynamic}, and integrating cost-sensitive thoughts into model training~\cite{sun2007cost}. 
However, during FL, the communication between clients and the server is restricted to the gradients and the common model. For privacy concern, it is preferable that the server does not require clients to upload additional information about their local data~\cite{geyer2017differentially, hamm2016learning}. Thus, it is infeasible to gather the information of all local data and conduct an aggregated analysis globally. 
This makes the vast majority of imbalance solutions not applicable to FL.
There are several approaches that may be applied locally, without uploading data distribution to the server~\cite{huang2016learning, wang2017learning, mikolov2013distributed}. However, due to the mismatch between local data distributions and the global one, these approaches are not effective and may even impose negative side-effect on the global model. 
The work in~\cite{duan2019astraea} directly addresses class imbalance in FL, but it requires clients to upload their local data distribution, which may expose latent backdoor to attacks and lead to privacy leakage. Moreover, it requires placing a number of proxy servers, which increases the FL complexity and incurs more computation overhead.

In this work, we tackle the above challenges. We consider FL as a scheme that is always in training~\cite{mcmahan2016communication}. During FL, new data is constantly generated by the clients, and class imbalance could happen at any time. If such imbalance cannot be detected in time, it will induce the common model to the wrong direction in the early training phase, and thus poison the common model and deteriorate the performance. 
To detect the imbalance in FL timely and accurately, we propose to design a monitor that estimates the composition across classes during FL, and if a particular imbalanced composition appears continuously, the monitor will alert the administrator ($\mathcal{AD}$) to apply measures that can mitigate the negative impact.
Moreover, we develop a new loss function \textbf{Ratio Loss}, and compare our approach to existing loss-based imbalance state-of-the-art solutions: \textbf{CrossEntropy Loss}, \textbf{Focal Loss}~\cite{lin2017focal} and \textbf{GHMC Loss}~\cite{li2019gradient}. Note that these loss functions are for general class imbalance problems, and their basis is just the output results of forward feeding. We choose them for comparison as they can also address the imbalance issue in FL without posing threats to privacy.
%In our view, these methods are most likely helpful to address the imbalance issue in FL without imposing threats to privacy.

\begin{figure}
  \centering
  \includegraphics[width=1.\linewidth]{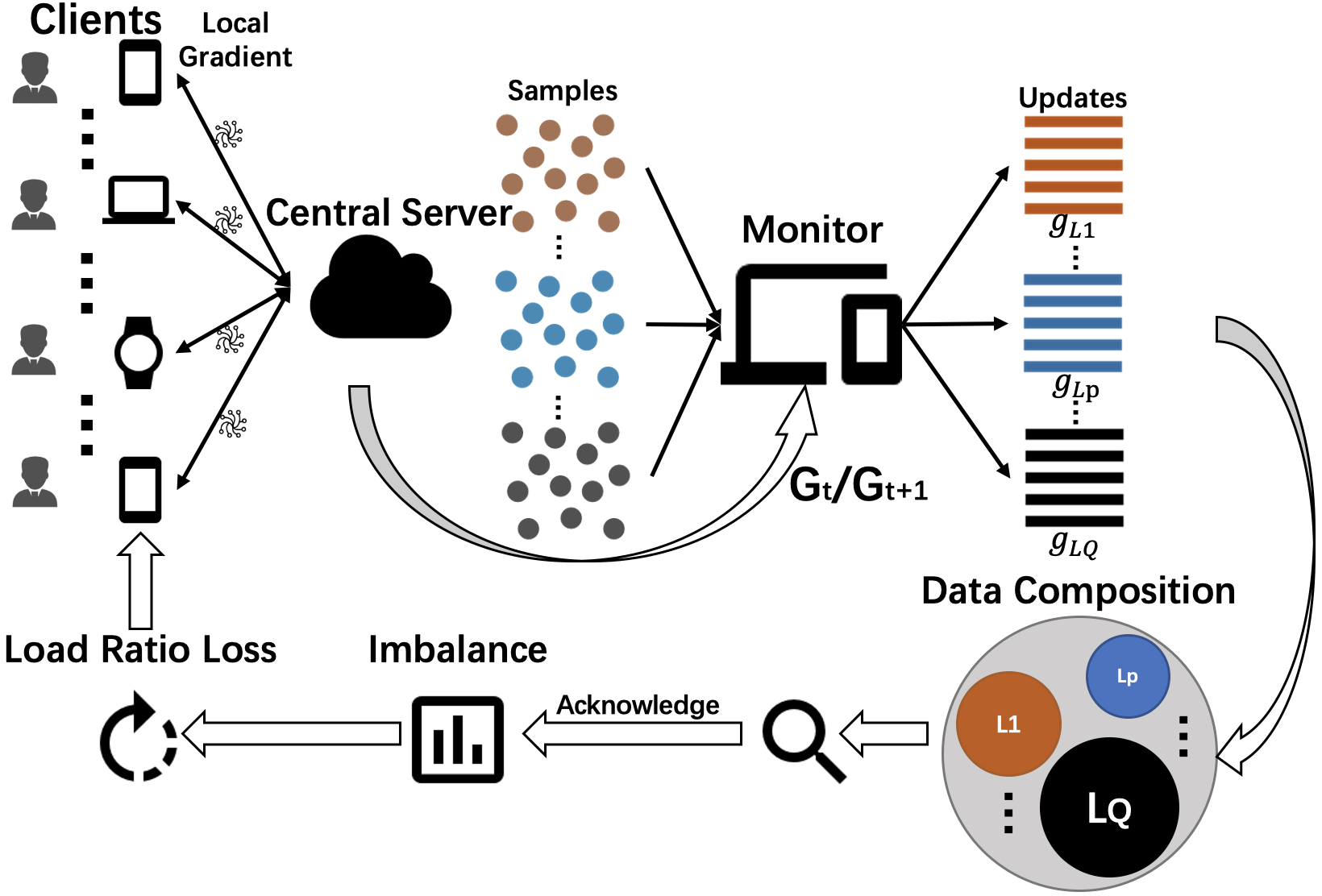}
  \caption{The monitor estimates the composition of training data round by round. When detecting a similar imbalanced composition continuously, the system will acknowledge the class imbalance and load the \textbf{Ratio Loss}.}
  \label{fig_workflow}
\end{figure}

The basic workflow of our method is shown in Fig.~\ref{fig_workflow}. At round $t\!+\!1$, the monitor downloads the global model $G_t$ of round $t$ and feeds samples of the auxiliary data to it. For each class, the monitor obtains corresponding gradient updates $g_L$. And by applying our method to compare these updates with $G_{t\!+\!1}$, our monitor can acquire the composition of training data at round $t\!+\!1$. If a similar imbalanced composition is detected continuously, the system will acknowledge that the global model has learned imbalanced data, and then try to mitigate its impact by applying the Ratio Loss in FL.

\smallskip
\noindent{\bf Our contributions.} More specifically, we made the following contributions in this work:
\begin{itemize}
    \item {Our approach monitors the composition of training data at each training round in a passive way. The monitor can be deployed at either the central server or a client device, and it will not incur significant computation overhead or impose threats to client privacy.}
    \item {Our works show the importance of acknowledging class imbalances as early as possible during FL and taking timely measures.}
    \item {Our approach defines two types of class imbalance in FL: \textbf{local imbalance} and \textbf{global imbalance}, and addresses class imbalance based on a new loss function (Ratio Loss). Our method is shown to significantly outperform previous state-of-the-art methods while maintaining privacy for the clients.}
\end{itemize}

%% file: Related_Work.tex
\section{Related Work}
\noindent{\bf Class Imbalance.}
In supervised learning, models require labeled training data for updating their parameters. The imbalance of the training data (i.e., the variation of the number of samples for different classes/labels) occurs in many scenarios, e.g., image recognition for disease diagnosis~\cite{xia2020cervical}, object detection~\cite{lin2017focal, wang2020single}.
Such class imbalance will worsen the performance of the learning models, especially decreasing the classification accuracy for minority classes~\cite{he2009learning}. Several works have designed new metrics~\cite{wang2016training} to quantify the model performance with class imbalance, rather than just considering the overall accuracy. Prior approaches to address class imbalance can be classified into three categories: data-level, algorithm-level, and hybrid methods. Data-level approaches leverage data re-sampling~\cite{van2007experimental, mani2003knn} and data augmentation~\cite{lee2016plankton, pouyanfar2018dynamic}. Algorithm-level approaches modify the training algorithm or the network structure, e.g., meta learning~\cite{ling2008cost, wang2020meta}, model tuning~\cite{pouyanfar2018dynamic}, cost-sensitive learning~\cite{cui2019class, wang2017learning}, and changing the loss function~\cite{lin2017focal, li2019gradient, luo2020grace}. Then, from the perspectives of both data and algorithm levels, hybrid methods emerge as a form of ensemble learning~\cite{liu2008exploratory, chawla2003smoteboost}.

As stated in the introduction, data-level methods cannot be applied in FL due to their violation of the privacy requirements. The cost-sensitive approaches need to analyze the distribution of training data, e.g., re-weighting the loss via inverse class frequency, and are not effective for FL due to the mismatch between local data distribution and the global one.
Other cost-sensitive methods need specific information of minority classes, e.g., \textbf{MFE Loss}~\cite{wang2016training} regards minority classes as positive classes, and calculates false positive and false negative to generate a new loss form. Such prior knowledge is also difficult to acquire in FL. To address class imbalance in FL, we believe that it is important to measure the imbalance according to the current common model rather than depending on the training data. We thus regard CrossEntropy Loss, Focal Loss~\cite{lin2017focal} and GHMC Loss~\cite{li2019gradient} as possibly methods to solve the class imbalance problem in FL, and compare our approach with them.

\smallskip
\noindent{\bf Federated Learning.} 
Due to the heavy computation burden for training deep learning models, researchers have been exploring using multiple devices to learn a common model. There are many studies on organizing multiple devices for distributed learning, with both centralized~\cite{kim2016deepspark} and decentralized~\cite{sergeev2018horovod} approaches. 
Recently, more and more local client devices (e.g, mobile phones) can participate in model learning.
Under such circumstances, the training data on local devices is more personal and privacy-sensitive. In order to avoid privacy leakage, federated learning~\cite{mcmahan2016communication, li2020federated} has emerged as a promising solution, which enables a global model to be learned while keeping all training data on local devices. The privacy protection in FL training is guaranteed by secure aggregation protocols~\cite{bonawitz2017practical} and differential privacy techniques~\cite{geyer2017differentially, niu2020billion}. In general, with these technologies, neither local participants nor the central server of FL can observe the individual gradient in the plain form during training. Despite of various types of inference attacks~\cite{melis2018exploiting, zhu2019deep, wang2019eavesdrop, sun2020data}, inferring information of particular clients is still extremely difficult. Therefore, how to extract useful information from the aggregated global gradient is interesting -- and in this work, we focus on extracting such information for addressing class imbalance.

%% file: Method.tex
\section{Methodology}
\subsection{Definition and Background}
Our problem is formulated on a multi-layer feed-forward neural network. Here we consider a classifier with output size equal to the number of classes $Q$. It is defined over a feature space $\mathcal{X}$ and a label space $\mathcal{Y} \!=\! \{1, \cdots, Q\}$. Without losing generality for our problem, we combine all the middle layers as a hidden layer $H\!L$. If we feed the $j$-th sample of the class $p$, denoted as $X_j^{(p)}$, to the classifier, its corresponding output of $H\!L$ is denoted as $Y_j^{(p)} \!=\! [y_{j,(1)}^{(p)}, ..., y_{j,(s)}^{(p)}]$. The output of the last layer is $Z_j^{(p)} \!=\! [z_{j,(1)}^{(p)}, ..., z_{j, (Q)}^{(p)}]$, followed by a softmax operation to obtain the probability vector $\mathcal{S} \!=\! [f_{j,(1)}^{(p)}, ..., f_{j,(Q)}^{(p)}]$. The $H\!L$ contains $s$ neurons. A function $f \!:\! \{\mathcal{X}\Rightarrow \mathcal{S}\}$ maps $\mathcal{X}$ to the output probability simplex $\mathcal{S}$, with $f$ parameterizing over the hypothesis class $\mathbb{W}$, i.e., the overall parameters of the classifier. Further, the connection weight from the $H\!L$ to the output layer is denoted as $\mathcal{W} \!=\! [\mathcal{W}_{(1)}, \mathcal{W}_{(2)}, ..., \mathcal{W}_{(Q)}]$, and $\mathcal{W} \!\in\! \mathbb{W}$. At each training iteration, we apply back-propagation to compute the gradient of loss $L(\mathbb{W})$ subject to $\mathbb{W}$. We use $\mathbb{W}(t)$ to denote the weights in the $t$-th training iteration, and $\lambda$ to denote the learning rate. We then have $\mathbb{W}(t\!+\!1) \!=\! \mathbb{W}(t) - \lambda \nabla L(\mathbb{W}(t))$.

\subsection{Monitoring Scheme}
We define two types of class imbalance in FL: \textbf{local imbalance} and \textbf{global imbalance}. On every local client device $j$, the number of samples for each class $p$, denoted by $N_p^j$, may vary. The local imbalance measures the extent of such variation on each client device. Specifically, we define the local imbalance $\gamma_j$ for device $j$ as the ratio between the sample number of the majority class on $j$ and the sample number of the minority class on $j$, i.e., $\gamma_j \!=\! \max_p\{N_p^j\} / \min_p\{N_p^j\}$, similar to the prevailing imbalance ratio measurement as in~\cite{buda2018systematic}. It is possible that $\min_p\{N_p^j\}\!=\!0$. We regard such situation as extreme imbalance, and consider them in our experiments. %are conducted with the consideration of extreme local imbalance situations.

From the global perspective, 
%considering the aggregation of all the training data on local devices, 
we can measure the extent of global class imbalance $\Gamma$ by defining it as the ratio between the total sample number of the majority class across all devices and that of the minority class, i.e.,  $\Gamma = \max_p\{\sum_j N_p^j\} / \min_p\{\sum_j N_p^j\}$.

In general, the local imbalance on each device may be different from the global imbalance, and in practice such difference could be quite significant. We may even encounter the cases where a particular class is the majority class on certain local devices but the minority class globally.
To better quantify such mismatch between local and global imbalance, we use a vector $v_j \!=\! [N_1^j, ..., N_Q^j]$ to denote the composition of local data on device $j$, where $Q$ is the overall number of classes; and we use another vector $V \!=\! [\sum_j N_1^j, ..., \sum_j N_Q^j]$ to denote the composition of global data. We then use cosine similarity (CS) score to compare their similarity, i.e., $C\!S_j = (v_j \cdot V) / (\|v_j\|\|V\|)$. 

In regular training scenarios, there is no distinction between local and global imbalance levels since the training data is centralized and accessible, and mitigating the negative impact of imbalance is much easier than in FL. Note that in FL, the local training can be regarded as regular centralized learning. Intuitively, we could utilize existing methods to address the local imbalance issue at every round locally. However, local models exist temporarily on the selected devices. They will not be applied for users' tasks and will be replaced with the latest global model at the next round.
As the result, addressing local imbalance may not have significant impact in FL. %, and we will pay attention to solving the problem of global imbalance. 
Moreover, because of the mismatch between local and global imbalance, simply adopting existing approaches at local devices is typically not effective and may even impose negative impact on the global model. Thus, we focus on addressing global imbalance in our work. To detect and mitigate the performance degradation caused by global imbalance, we develop a monitoring scheme to estimate the composition of training data during FL, as explained below. 

\smallskip
\noindent{\bf Proportional Relation.}
We will first analyze the relation between the gradient magnitude and the sample quantity.

\smallskip
\noindent \textbf{Theorem 1}: \emph{For any real-valued neural network $f$ whose last layer is a linear layer with a softmax operation (without any parameter sharing), and for any input sample $X_i^{(p)}$ and $X_j^{(p)}$ of the same class $p$, if the inputs of the last layer $Y_i$ and $Y_j$ are identical, the gradients of link weights $\mathcal{W}$ between the last layer and its former layer induced by $X_i^{(p)}$ and $X_j^{(p)}$ are identical.}

The proof of Theorem 1 is presented in the Supplementary Materials ($\mathcal{S\!M}$). In the mini-batch training, gradients of samples within the batch are accumulated to update the model parameters, i.e., 
\begin{equation}
    \Delta_{\text{batch}}\mathcal{W} = - \frac{\lambda}{n^{\text{batch}}} \sum_{p=1}^{Q} \sum_{j=1}^{n^{(p)}} \nabla_{\mathcal{W}_j^{(p)}} L(\mathbb{W})
    \label{eq_batch_gradient}
\end{equation}

From our empirical study (please see $\mathcal{S\!M}$), we observe that the data samples of a same class $p$ induce similar $Y^{(p)}$s, and thus their corresponding gradients are very similar. If the average value of the gradients is $\overline{\nabla_{\mathcal{W}^{(p)}} L(\mathbb{W})}$,  Eq.~(\ref{eq_batch_gradient}) can be written as:
\begin{equation}
    \Delta_{\text{batch}}\mathcal{W} = - \frac{\lambda}{n^{\text{batch}}} \sum_{p=1}^{Q} \left (\overline{\nabla_{\mathcal{W}^{(p)}} L(\mathbb{W})} \cdot n^{(p)} \right )
\end{equation}
where $n^{(p)}$ is the number of samples for class $p$ in this batch, and $n^{\text{batch}}$ is the batch size. For one round of local training in FL, the total iteration number of local gradient update is $\left [\left( \sum_{p=1}^{Q} N_p / n^{\text{batch}}\right ) \cdot \text{N}_{ep} \right ]$, where $\text{N}_{ep}$ denotes the number of local epochs. % \textcolor{red}{this formula is not very clear; `number of local epochs'? Is it on denominator or numerator?}. 
To illustrate the proportional relation between gradient magnitude and sample quantity, we assume that the parameter change is relatively small and can be neglected within a training epoch. In this case, $\overline{\nabla_{\mathcal{W}^{(p)}} L(\mathbb{W})}$ of different batches within an epoch remains the same, and we can aggregate them and obtain the weight update of one epoch as:

\begin{equation}
    \Delta_{\text{epoch}}\mathcal{W} = - \frac{\lambda}{n^{\text{batch}}} \sum_{p=1}^{Q} \left (\overline{\nabla_{\mathcal{W}^{(p)}} L(\mathbb{W})} \cdot N_p \right )
\end{equation}
where $N_p$ is the overall sample number of class $p$. 

In the setting of standard FL, the selected local gradients are aggregated by the FedAvg~\cite{mcmahan2016communication}:
\begin{equation}
    \nabla L(\mathbb{W})_{t\!+\!1}^{A\!v\!g} = \frac{1}{K}\sum_{j=1}^K\nabla L(\mathbb{W})_{t\!+\!1}^{j}
    \label{FedAvg}
\end{equation}
where $K$ represents the number of selected clients. In this work, we consider the case where feature spaces of data sets owned by different clients are similar~\cite{yang2019federated}. %there are differences of feature space among data sets owned by clients, but these differences cannot be presented great. 
In the case where they have significant differences, transfer learning techniques such as domain adaptation~\cite{ganin2015unsupervised, dong2020can} may be needed to reduce the distribution discrepancy of different clients. This can be viewed as a problem of Federated Transfer Learning (FTL) with feature and model heterogeneity, and we plan to investigate the imbalance problem of FTL in our future work. 
%Otherwise, it is essential that incorporating transfer learning techniques, e.g., domain adaptation, to shrink the distribution discrepancy of different clients, which is the problem of Federated Transfer Learning (FTL) -- feature and model heterogeneity (we will investigate the imbalance problem of FTL in our future work). 

Based on above analysis, for any local training starting from the same current global model, data samples of the same class $p$ output very similar $Y$ (see $\mathcal{S\!M}$) and similar $\overline{\nabla_{\mathcal{W}^{(p)}} L(\mathbb{W})}$. In this case, the gradient induced by class $p$ in one global epoch is:
\begin{equation}
\begin{split}
    \Delta_{\text{global}}\mathcal{W}^{(p)} &= - \frac{\lambda}{n^{\text{batch}} \cdot K} \sum_{j=1}^{K} \left (\overline{\nabla_{\mathcal{W}^{(p)}}^j L(\mathbb{W})} \cdot N_p^j \right ) \\
    &= - \frac{\lambda}{n^{\text{batch}} \cdot K} \cdot \overline{\nabla_{\mathcal{W}^{(p)}} L(\mathbb{W})} \left (\sum_{j=1}^K N_p^j \right )
\end{split}
\end{equation}

Based on this relation, we develop our monitoring scheme as follows. In round $t\!+\!1$, the monitor will feed samples of every class in the auxiliary data singly to the same global model of round $t$, i.e., $G_t$. It then obtains corresponding weight updates $\{g_{L_1}, ..., g_{L_p}, ..., g_{L_Q}\}$, where each $g_{L_p}$ corresponds to the class $p$. In practice, we observe that not all weights get updated significantly -- some of them increase little and thus easily get offset by the negative updates of other classes. Accordingly, we design a filter to select the weights whose updating magnitudes are relatively large. Specifically, for class $p$, we get the weight updates of the $p$-th output node  (denoted as $\Delta\mathcal{W}_{(p)}^{(1 \sim Q)}$) from $\{g_{L_1}, ..., g_{L_p}, ..., g_{L_Q}\}$, and compute the ratio $Ra_{p, i}$ for each weight component of $\Delta \mathcal{W}_{(p)}$ as follows:
\begin{equation}
    Ra_{p, i} = \frac{(Q - 1) \Delta \mathcal{W}_{(p, i)}^{(p)}}{\sum_{j=1}^Q(\Delta\mathcal{W}_{(p, i)}^{(j)}) - \Delta\mathcal{W}_{(p, i)}^{(p)}}
    \label{ratio}
\end{equation} 
where $i\!=\!1,...,s$. We set a threshold $T_{Ra}$ ($T_{Ra}\!=\!1.25$; refer to $\mathcal{S\!M}$ for the experiments of setting $T_{Ra}$), and we select components of $\Delta \mathcal{W}_{(p)}$ whose ratios $Ra_{p, i}$ are larger than $T_{Ra}$. Based on the proportional relation, we formulate the accumulation of weight changes under FedAvg:
\begin{equation}
\begin{split}
    \Delta\mathcal{W}_{(p, i)}^{(p)}\cdot \widehat{N}_{p, i} &+ \left(\sum_{j=1}^K \! \sum_{p=1}^{Q} N_p^j - \widehat{N}_{p, i}\right) \frac{\Delta \mathcal{W}_{(p,i)}^{(p)}}{Ra_{p, i}} \\
    &= n^p_a \cdot K\left(\mathcal{W}_{(p, i)}^{G_{t\!+\!1}} - \mathcal{W}_{(p, i)}^{G_t}\right)
    \label{number calculation}
\end{split}
\end{equation}
where $n^p_a$ is the sample number of class $p$ in the auxiliary data, $\widehat{N}_p$ is the predicted sample quantity of class $p$, $\mathcal{W}_{(p, i)}^{G_t}$ and $\mathcal{W}_{(p, i)}^{G_{t\!+\!1}}$ are link weights $\mathcal{W}_{(p, i)}$ of $G_t$ and $G_{t\!+\!1}$, respectively. $\sum_{p=1}^Q N_p^j$ is the overall number of all samples owned by client $j$, and we need clients to upload $\sum_{p=1}^Q N_p^j$ to the server. This is the only information needed from clients in our monitoring scheme. We believe that this is a reasonable assumption of a trade-off between client privacy and the system’s capability to estimate class distribution. First, sharing the sample quantity of each class may lead to serious privacy concerns, which is essential for other imbalance solutions. Malicious attackers that obtain such information could analyze client preferences (e.g., what type of image/music/search words each client accesses more), group them based on class distributions, and send targeted information or launch targeted attacks. Besides, the class distribution itself is an attack surface as shown in~\cite{salem2020updates}, which proposes methods to reconstruct the training data by leveraging class distribution. However, sharing only the total sample quantity across all classes carries a much lower risk and can be protected by secure aggregation, e.g., methods in~\cite{salem2020updates} cannot launch reconstruction attacks anymore, and it is hard for attackers to send meaningful targeted information as well. Thus, we believe that sharing the total sample quantity is a reasonably small privacy cost that we can pay to effectively address FL class imbalance.

Now, except for $\widehat{N}_{p, i}$, all values in Eq.~(\ref{number calculation}) can be acquired by the monitor. We can then use Eq.~(\ref{number calculation}) to compute $\widehat{N}_{p, i}$ for each component of the filtered $\Delta \mathcal{W}_{(p)}$.
And we can obtain the final result as the average value of all calculated $\widehat{N}_{p, i}$ (denoted as $\widehat{N}_p$). After the computation for all classes, we can obtain the proportion vector of the current training round $v_{pt}=[\widehat{N}_1, ..., \widehat{N}_p, ..., \widehat{N}_Q]$, an estimation of the data composition of the current round. 

\subsection{Ratio Loss based Mitigation}
Once our monitor detects a similar imbalanced composition continuously by checking $v_{pt}$, it will acknowledge that the global model has learned imbalanced data and apply a mitigation strategy that is based on our \textbf{Ratio Loss} function.

As aforementioned, applying existing approaches locally will not be effective in mitigating the impact of global imbalance. Our method instead measures the global imbalance based on the current global model. According to previous analysis, weight updates are proportional to the quantity of samples for different classes, and the current network is built by accumulating such updates round by round. Due to the difference of feature space among classes, it is likely more reasonable to use the contribution to gradient updates rather than just the number of data samples for demonstrating the imbalance problem. In other words, after feeding some data to train the network, if weights of different nodes are updated similarly in terms of magnitude, we can regard the training as balanced, and vice versa. Because the layers before output nodes are shared by all classes, we restrict our interest on link weights between the $H\!L$ and output nodes. Specifically, we consider the imbalance problem in FL as \textit{the weight changes of different output nodes present noticeable magnitude gap when feeding corresponding classes}.

\smallskip
\noindent \textbf{Theorem 2}: \emph{For any real-valued neural network $f$ whose last layer is a linear layer with a softmax operation (without any parameter sharing), and the activation function between the last layer and its former layer is non-negative (e.g., Sigmoid and ReLu), if $f$ has learned imbalanced data set, for any majority class $\mathcal{A}$, any minority class $\mathcal{B}$, and another class $\mathcal{C}$ ($\mathcal{C} \neq \mathcal{A} \,\text{and}\, \mathcal{C} \neq \mathcal{B}$, but $\mathcal{C}$ can be any class other than chosen $\mathcal{A}\, \&\, \mathcal{B}$) fed to $f$, we have:}
\begin{equation}
    |\nabla_{\mathcal{W}_{(\mathcal{A})}^{(\mathcal{C})}} L(\mathbb{W})| > |\nabla_{\mathcal{W}_{(\mathcal{B})}^{(\mathcal{C})}} L(\mathbb{W})|
\end{equation}

\noindent \textbf{Assumption}: 1) The input similarity between class $\mathcal{C}$ and $\mathcal{A}$ is the same as between class $\mathcal{C}$ and $\mathcal{B}$. 2) The reason why there is classification accuracy degradation on the minority class $\mathcal{B}$ is that its probability result $f_{(\mathcal{B})}^{(\mathcal{B})}$ is not distinguishable with the output of other $f_{(i)}^{(\mathcal{B})}$ ($i \!=\! 1,...,Q \, \mbox{and} \, i \neq \mathcal{B}$). Thus minority classes can be regarded as hard samples generally, while majority classes are easy samples, i.e., $\left(f_{(\mathcal{A})}^{(\mathcal{A})} / f_{(\mathcal{B})}^{(\mathcal{A})}\right) \!\gg\! \left(f_{(\mathcal{B})}^{(\mathcal{B})} / f_{(\mathcal{A})}^{(\mathcal{B})}\right) \!>\! 1$.

The detailed proof is shown in the $\mathcal{S\!M}$. Based on Theorem 2, we propose to mitigate global imbalance by designing our \textbf{Ratio Loss} function, denoted as $L_{R\!L}$. Specifically, we first consider the widely-used \textbf{CrossEntropy Loss} function (denoted as $L_{C\!E}$) for the multi-class classifier:
\begin{equation}
    L_{C\!E} = - p \cdot [log(\mathcal{S})]
\end{equation}
where $p$ is the ground-truth label and always the one-hot form in multi-class classifiers, while $\mathcal{S}$ denotes the vector of probability results. In order to address imbalance, a common method is to introduce a weight vector $\Pi = [\pi_1, ..., \pi_Q]$ and there is $\Pi \cdot L_{C\!E}$. Typically, $\pi$ is determined by the proportions or frequencies of different classes for the overall training data. Intuitively, a larger proportion corresponds to a lower $\pi$, and vice versa. 

As stated above, we use the noticeable differences of weight changes to evaluate the global imbalance. Taking $L_{C\!E}$ as the basic term, we define the Ratio Loss $L_{R\!L}$ as:
\begin{equation}
    L_{R\!L} = (\alpha + \beta \mathbb{R}) \cdot p \cdot L_{C\!E}
\end{equation}
where $\alpha$ and $\beta$ are two hyper-parameters. In our experiments, when $\alpha\!=\!1.0$ and $\beta\!=\!0.1$, the mitigation results are the best (the experiments for setting $\alpha$ and $\beta$ can be found in the $\mathcal{S\!M}$). After computing all $Ra_{p, i}$ for class $p$ as Eq.~(\ref{ratio}), we can get their average value and its corresponding absolute value, denoted as $Ra_{p}$, and compose $\mathbb{R} = [Ra_1, ..., Ra_p, ..., Ra_Q]$. Finally, in the local training, when a sample $X^{(p)}$ of class $p$ is fed to the neural network, its corresponding loss is:
\begin{equation}
    L_{R\!L}(X^{(p)}) = -(\alpha + \beta Ra_p) \cdot log(f_{(p)}^{(p)})
\end{equation}

We mitigate the impact of class imbalance by modifying the coefficient $\pi$ before $L_{C\!E}$. When the input is a minority class, according to Theorem 2, its corresponding $Ra$ is relatively large, and then its contribution to the overall loss will increase, and vice versa. Compared with \textbf{GHMC Loss}, \textbf{Ratio Loss} pays attention to the gradient on the output node corresponding to the ground-truth label of data samples, and also considers the impact over gradients on the same node from samples of other classes. In addition, the utilization of $L_{R\!L}$ does not require clients to upload their overall sample quantities -- $\sum_{p=1}^Q N_p^j$, which maintains the client privacy. 

%% file: Evaluation.tex
\section{Experimental Results}
\subsection{Experiment Setup}
 We implement the algorithms mainly in PyTorch.
Our experiments follow the standard structure of FL~\cite{konevcny2016federated, mcmahan2016communication}. We choose four different datasets: MNIST, CIFAR10, Fer2013~\cite{goodfellow2013challenges}, and FEMNIST of LEAF benchmark~\cite{caldas2018leaf}. Fer2013 relates to face recognition and has imbalance issue, and FEMNIST is a popular FL data set with great feature heterogeneity. For each data set, we utilize the following convolution neural networks: LeNet5 for MNIST, a 7-layer CNN for CIFAR10, Resnet18~\cite{he2016deep} for Fer2013, and Resnet50 for FEMNIST.
The local training batch size is $32$, the learning rate $\lambda \!=\! 0.001$, and the optimizer is SGD.
The auxiliary data is a set of samples of different classes that is fed into the current global model by the monitor. 
It can be acquired from the public data or be synthesized by the FL administrator who has legal access to prior knowledge about what the training data may look like. Such auxiliary data can be utilized for a long time, unless the training data of the overall FL system changes significantly. Moreover, the required size of the auxiliary data is small. In our experiments, we use only $32$ samples for each class, while the sample quantity of a client is more than $10,000$. Due to the small size of the auxiliary data, the deployment of the monitor does not incur significant computation overhead, and we indeed did not observe noticeable additional processing time during our experiments. Please refer to the $\mathcal{S\!M}$ for more details about the auxiliary data, and the setting for hardware.
%For hardware setting, please see our $\mathcal{S\!M}$.

\subsection{Effectiveness of Monitoring Scheme}
\label{monitor experiment}
\begin{figure}[htbp]
\centering
\includegraphics[width=.75\linewidth]{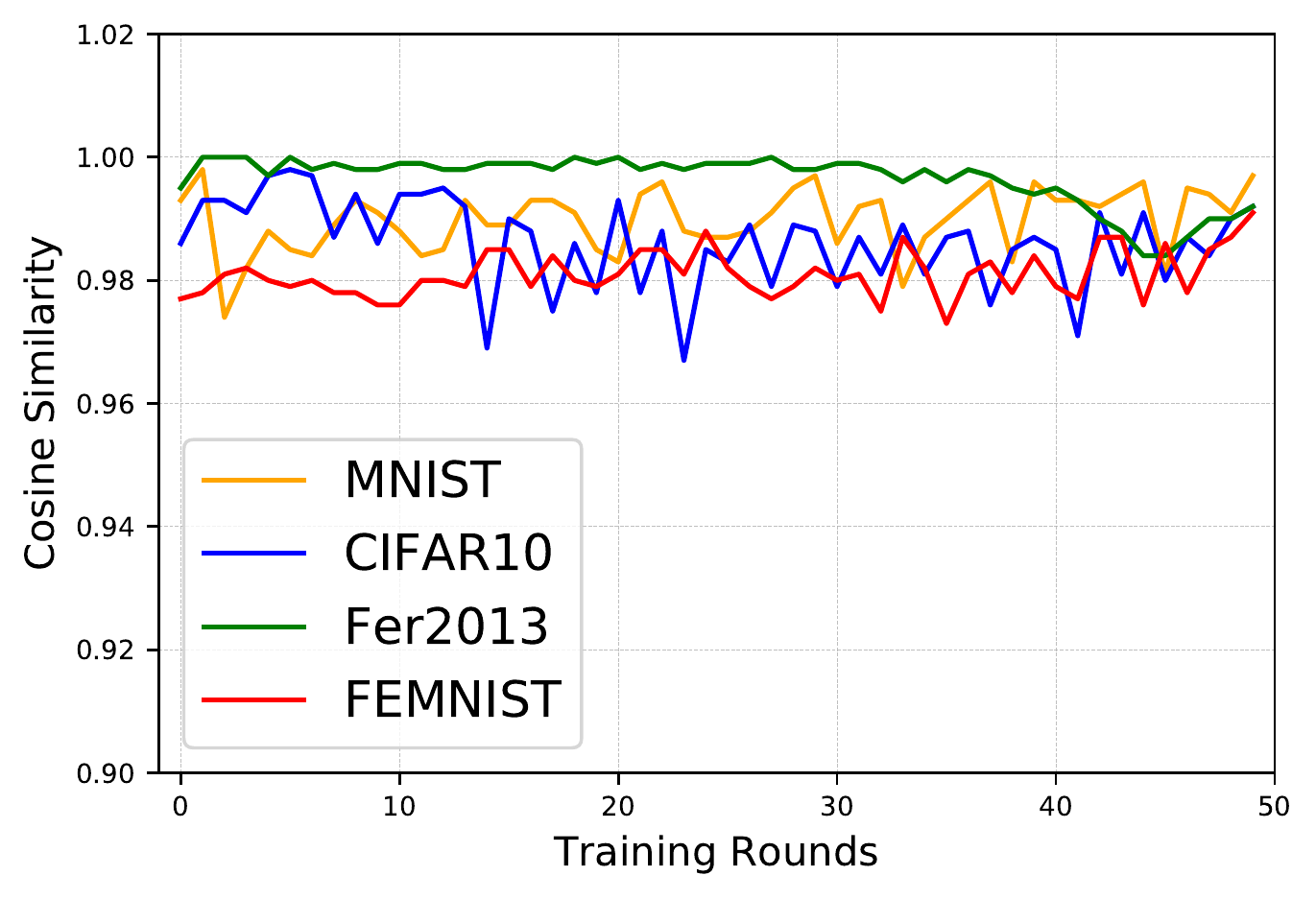}
\caption{Similarity between our estimation of the global data composition and the ground truth.}
\label{fig_monitoring}
\end{figure}

To evaluate the effectiveness of our monitoring scheme, we design the experiments as the central server randomly selects $20$ clients from $100$ participants during each round of the FL training. Each client trains its model for $10$ local epochs for MNIST and FEMNIST, and $5$ for CIFAR10 and Fer2013 (in this case, one global round for each data set costs nearly the same amount of time). Training $30$ global rounds can make the model on MNIST converge, $50$ for CIFAR10, Fer2013, and FEMNIST. 
For each client, first the number of classes they have locally is randomly determined as an integer between $1$ and $Q$.  Then, the specific classes are randomly chosen for each client, with equal sample quantity for each class. For FEMNIST, as each writer has a relatively small number of data samples (several dozens), we group 20 writers into a new client and thus turn approximately 2,000 writers into 100 clients (we believe that the heterogeneity holds true under this allocation strategy). For all data sets, we allocate them to clients without replacement, and the detailed data splitting is visualized in the $\mathcal{S\!M}$. During FL, different client selections at each round lead to varying data compositions, and thus different global imbalance patterns and various non-IID situations. 
As introduced, our monitor computes a data composition vector $v_{pt}$ for each training round. We can compare it against the ground truth, defined as $V$.
Fig.~\ref{fig_monitoring} shows the comparison between our estimated $v_{pt}$ and $V$, measured by a cosine similarity score. The closer it is to $1$, the more accurate our estimation is. From the figure, we can observe that our estimation of the data composition is very close to the ground truth. Among four datasets, the average similarity score is above $0.98$ and higher than $0.99$ for most of the time. Such results demonstrate the effectiveness of our monitoring scheme. We also carry out experiments with different numbers of clients, and we find that the similarity score gets even closer to $1$ with the increase of client number. We also find that the local batch size and epochs have little impact on the performance of the monitoring scheme. The detailed results are in the $\mathcal{S\!M}$.

\begin{table}[htbp]
\centering
\small
\setlength{\tabcolsep}{1.9mm}{
\begin{tabular}{c|c|ccccc}
\hline
\multicolumn{2}{c|}{\textbf{Data}}                                              & \multicolumn{5}{c}{\textbf{FEMNIST}}                                               \\ \hline
\multicolumn{2}{c|}{$\Gamma$}                                                & B.             & 10:1           & 20:1           & 50:1           & 100:1          \\ \hline
\multirow{4}{*}{\textbf{\begin{tabular}[c]{@{}c@{}}Ac.M\\ \%\end{tabular}}} & $L_{C\!E}$ & 90.46          & 74.32          & 62.64          & 33.25          & 18.48          \\
                                                                            & $L_{F\!L}$ & 91.25          & 75.96          & 64.14          & 38.16          & 25.11          \\
                                                                            & $L_{G\!H\!M\!C}$ & 92.64          & 79.75          & 69.29          & 42.55          & 29.16          \\
                                                                            & $L_{R\!L}$(ours) & \textbf{93.46} & \textbf{88.48} & \textbf{72.29} & \textbf{51.66} & \textbf{41.45} \\ \hline
\multirow{4}{*}{\textbf{AUC}}                                               & $L_{C\!E}$ & .9652          & .9441          & .9187          & .8979          & .8667          \\
                                                                            & $L_{F\!L}$ & .9650          & .9540          & .9291          & .9011          & .8774          \\
                                                                            & $L_{G\!H\!M\!C}$ & .9691          & .9542          & .9393          & .9023          & .8842          \\
                                                                            & $L_{R\!L}$(ours) & \textbf{.9699} & \textbf{.9607} & \textbf{.9477} & \textbf{.9138} & \textbf{.9001} \\ \hline
\end{tabular}
}
\caption{Comparison between our method ($L_{R\!L}$) and previous methods based on CrossEntropy Loss ($L_{C\!E}$), Focal Loss ($L_{F\!L}$) and GHMC Loss ($L_{G\!H\!M\!C}$) in federate learning, over FEMNIST and different levels of global imbalance.}
\label{FL_FEMNIST}
\end{table}

\begin{table*}[htb]
\centering
\small
\setlength{\tabcolsep}{1.45mm}{
\begin{tabular}{c|c|ccccc|ccccc|ccccc}
\hline
\multicolumn{2}{c|}{\textbf{Data}}                                              & \multicolumn{5}{c|}{\textbf{MNIST}}                                                & \multicolumn{5}{c|}{\textbf{CIFAR10}}                                              & \multicolumn{5}{c}{\textbf{Fer2013}}                                               \\ \hline
\multicolumn{2}{c|}{$\Gamma$}                                             & B.             & 10:1           & 20:1           & 50:1           & 100:1          & B.             & 10:1           & 20:1           & 50:1           & 100:1          & B.             & 10:1           & 20:1           & 50:1           & 100:1          \\ \hline
\multirow{4}{*}{\textbf{\begin{tabular}[c]{@{}c@{}}Ac.M\\ \%\end{tabular}}} & $L_{C\!E}$ & \textbf{98.22} & 90.19          & 80.04          & 63.66          & 46.84          & 57.57          & 23.43          & 15.17          & 04.93          & 00.97          & \textbf{97.93} & 23.59          & 12.65          & 05.43          & 02.41          \\
                                                                            & $L_{F\!L}$ & 96.42          & 84.84          & 75.65          & 63.43          & 41.76          & 50.10          & 26.40          & 17.77          & 06.47          & 01.57          & 85.28          & 21.87          & 12.86          & 05.56          & \textbf{03.01} \\
                                                                            & $L_{G\!H\!M\!C}$ & 93.05          & 81.24          & 64.98          & 61.38          & 20.23          & 50.10          & 27.73          & 19.13          & \textbf{06.77} & 02.53          & 46.54          & 19.76          & 08.72          & 02.44          & 01.47          \\
                                                                            & $L_{R\!L}$(ours) & 98.04          & \textbf{92.05} & \textbf{81.70} & \textbf{74.51} & \textbf{56.50} & \textbf{63.23} & \textbf{29.77} & \textbf{19.17} & \textbf{06.77} & \textbf{03.03} & 97.87          & \textbf{25.55} & \textbf{13.34} & \textbf{06.46} & 02.95          \\ \hline
\multirow{4}{*}{\textbf{AUC}}                                               & $L_{C\!E}$ & .9907          & .9780          & .9526          & .9338          & .9056          & .7425          & .6944          & .6777          & .6628          & .6578          & \textbf{.9893} & .7932          & \textbf{.7574} & .7320          & \textbf{.7275}          \\
                                                                            & $L_{F\!L}$ & .9830          & .9642          & .9485          & .9282          & .8927          & .7028          & .6790          & .6691          & .6498          & .6584          & .9473          & .7599          & .7337          & .7241          & .7184          \\
                                                                            & $L_{G\!H\!M\!C}$ & .9620          & .9461          & .9216          & .9184          & .8419          & .7197          & .6945          & .6916          & .6735          & .6629          & .8271          & .7696          & .7429          & .7081          & .7074          \\
                                                                            & $L_{R\!L}$(ours) & \textbf{.9908} & \textbf{.9815} & \textbf{.9644} & \textbf{.9531} & \textbf{.9213} & \textbf{.7678} & \textbf{.7179} & \textbf{.7084} & \textbf{.6844} & \textbf{.6820} & .9891          & \textbf{.7962} & .7482          & \textbf{.7372} & .7268 \\ \hline
\end{tabular}
}
\caption{Comparison between our method ($L_{R\!L}$) and previous methods based on CrossEntropy Loss ($L_{C\!E}$), Focal Loss ($L_{F\!L}$) and GHMC Loss ($L_{G\!H\!M\!C}$) in federate learning, over three data sets and different levels of global imbalance.}
\label{Res_3_datasets}
\end{table*}

\begin{table*}[htb]
\centering
\small
\setlength{\tabcolsep}{1.45mm}{
\begin{tabular}{c|c|ccccc|ccccc|ccccc}
\hline
\multicolumn{2}{c|}{\textbf{Data}}                                              & \multicolumn{5}{c|}{\textbf{MNIST}}                                                & \multicolumn{5}{c|}{\textbf{CIFAR10}}                                              & \multicolumn{5}{c}{\textbf{Fer2013}}                                               \\ \hline
\multicolumn{2}{c|}{$\gamma$}                                                & B.             & 10:1           & 20:1           & 50:1           & 100:1          & B.             & 10:1           & 20:1           & 50:1           & 100:1          & B.             & 10:1           & 20:1           & 50:1           & 100:1          \\ \hline
\multirow{4}{*}{\textbf{\begin{tabular}[c]{@{}c@{}}Ac.M\\ \%\end{tabular}}} & $L_{C\!E}$ & 98.68          & 90.14          & 85.86          & 75.64          & 51.20          & 73.07          & 17.93          & 11.84          & 01.53          & 00.47          & 83.23          & 10.95          & 04.70          & 01.85          & 00.91          \\
                                                                            & $L_{F\!L}$ & \textbf{99.22} & 93.88          & 87.03          & 78.70          & 58.55          & 65.40          & 27.90          & 16.00          & 05.70          & 02.37          & 83.04          & 14.07          & \textbf{08.18}          & 02.31          & 00.83          \\
                                                                            & $L_{G\!H\!M\!C}$ & 97.55          & 92,91          & 88.03          & 78.84          & 59.01          & 74.07          & 28.10          & 17.04          & 05.87          & 02.35          & 83.31          & 14.32          & 06.01          & 01.85          & 00.86          \\
                                                                            & $L_{R\!L}$(ours) & 98.59          & \textbf{93.99} & \textbf{89.85} & \textbf{79.41} & \textbf{60.34} & \textbf{76.33} & \textbf{29.87} & \textbf{17.70} & \textbf{06.43} & \textbf{02.40} & \textbf{83.36} & \textbf{14.93} & 06.99 & \textbf{02.46} & \textbf{00.93} \\ \hline
\multirow{4}{*}{\textbf{AUC}}                                               & $L_{C\!E}$ & .9929          & .9793          & .9729          & .9543          & .9108          & .8429          & .7354          & .7183          & .7078          & .7068          & .8831          & .6975          & .6853          & .6752          & .6745          \\
                                                                            & $L_{F\!L}$ & \textbf{.9934} & .9862          & .9739          & .9612          & .9151          & .8001          & .7689          & .7530          & .7442          & .7318          & .8713          & .7135          & .7029          & .6894          & .6893          \\
                                                                            & $L_{G\!H\!M\!C}$ & .9895          & .9862          & .9799          & .9615          & .9237          & .8470          & .7773          & .7658          & .7444          & .7309          & .8881          & .7233          & .7030          & \textbf{.7041} & .6902          \\
                                                                            & $L_{R\!L}$(ours) & .9932          & \textbf{.9864} & \textbf{.9801} & \textbf{.9625} & \textbf{.9306} & \textbf{.8624} & \textbf{.7868} & \textbf{.7712} & \textbf{.7447} & \textbf{.7416} & \textbf{.8883} & \textbf{.7236} & \textbf{.7049} & .6927          & \textbf{.6905} \\ \hline
\end{tabular}
}
\caption{Comparison between our method ($L_{R\!L}$) and previous methods based on CrossEntropy Loss ($L_{C\!E}$), Focal Loss ($L_{F\!L}$) and GHMC Loss ($L_{G\!H\!M\!C}$), when the models are not trained with federate learning.}
\label{Res_without_mismatch}
\end{table*}

\begin{table*}[h!]
\centering
\small
\setlength{\tabcolsep}{1.99mm}{
\begin{tabular}{c|c|cccc|cccc|cccc}
\hline
\multicolumn{2}{c|}{\textbf{Data}}                                              & \multicolumn{4}{c|}{\textbf{MNIST}}        & \multicolumn{4}{c|}{\textbf{CIFAR10}}     & \multicolumn{4}{c}{\textbf{Fer2013}}       \\ \hline
\multicolumn{2}{c|}{$C\!S$}                                                & 0.6960          & 0.6158 & 0.5111 & 0.3984 & 0.6960         & 0.6158 & 0.5111 & 0.3984 & 0.9343 & 0.8489 & 0.7411 & 0.6732 \\ \hline
\multirow{2}{*}{\textbf{\begin{tabular}[c]{@{}c@{}}Ac.M\\ \%\end{tabular}}} & $L_{M\!S\!E}$ & 70.36           & 60.88  & 45.68  & 00.60  & 01.87          & 01.40  & 01.70  & 07.60  & 48.77  & 37.25  & 18.25  & 03.55  \\
                                                                            & $L_{M\!F\!E}$ & \textbf{71.33}  & 59.71  & 40.42  & 00.00  & \textbf{01.97} & 01.37  & 01.60  & 07.03  & 46.41  & 35.01  & 16.47  & 03.14  \\ \hline
\multirow{2}{*}{\textbf{AUC}}                                               & $L_{M\!S\!E}$ & .9397          & .9214 & .8848 & .7775 & .6638         & .6450 & .6267 & .5842 & .8564 & .8259 & .7772 & .7400 \\
                                                                            & $L_{M\!F\!E}$ & \textbf{.9417} & .9167 & .8787 & .7754 & .6623         & .6448 & .6256 & .5820 & .8502 & .8209 & .7738 & .7382 \\ \hline
\end{tabular}
}
\caption{Comparison between the Mean-Square-Error Loss ($L_{M\!S\!E}$), and the MFE Loss with local knowledge in FL setting ($L_{M\!F\!E}$) over three data sets and under different levels of mismatch between local and global imbalance (measured by $C\!S$).} 
\label{Sup_MFE}
\end{table*}

\subsection{Overall Comparison with Previous Methods}
We then conduct experiments to evaluate the effectiveness of our Ratio Loss ($L_{R\!L}$), and compare it with CrossEntropy Loss ($L_{C\!E}$), Focal Loss ($L_{F\!L}$) and GHMC Loss ($L_{G\!H\!M\!C}$). We use the similar experiment setting as previous, except that we now explicitly explore different levels of global imbalance $\Gamma$, i.e., setting the ratio as $10\!:\!1$, $20\!:\!1$, $50\!:\!1$, and $100\!:\!1$, respectively, and we also include experiments when the data is balanced (B.). The evaluation metrics are the AUC score and the classification accuracy of minority classes (Ac.M). The results for majority classes can be found in the $\mathcal{S\!M}$. To keep $\Gamma$ unchanged during training, we fix the selected clients as those chosen in the first round. (We also conduct experiments when the $\Gamma$ is dynamically changing during FL training, and our $L_{R\!L}$ also performs the best. Please refer to our $\mathcal{S\!M}$ for results.)

After demonstrating the importance of early acknowledgment for global imbalance (please refer to our $\mathcal{S\!M}$), we quantitatively compare our method with previous ones in Ac.M and AUC score. The results are shown in Table~\ref{FL_FEMNIST} for FEMNIST and Table~\ref{Res_3_datasets} for MNIST, CIFAR10 and Fer 2013 (all data points are the average of 5 runs).  
%for each situation, we conduct experiments for 5 times, and we obtain the average performance shown in Table~\ref{FL_FEMNIST} and Table~\ref{Res_3_datasets}. 
We can see that our method can effectively mitigate the impact of class imbalance and outperform the previous methods in almost all cases. Our improvement is particularly significant for MNIST and FEMNIST. 
%, where the absolute improvement of Ac.M ranges between $6\%$-$15\%$. and for the imbalance level of $10\!:\!1$ across all data sets, where the absolute improvement of Ac.M ranges between $3\%$-$7\%$. And for FEMNIST, the improvement of Ac.M ranges between $10\%$-$23\%$.

Moreover, we also compare our method with previous ones for the regular training of neural networks \emph{without} FL. Table~\ref{Res_without_mismatch} demonstrates that in these cases, our method still outperforms the other three in most scenarios. This shows the broader potential of our Ratio Loss function.  

\subsection{Mismatch between Local and Global Imbalance}
We also conduct a set of experiments to explore the impact of the mismatch beween local and global imbalance. We adjust the mismatch level by setting different number of classes each client may have, i.e., from $2$ to $5$ out of a total number of $10$ classes globally. Intuitively, the smaller the number, the less representative each client is with respect to the global training set, and hence the larger the mismatch.

To start with, we implemented \textbf{MFE Loss}~\cite{wang2016training} in FL as the representative method aiming to address the local imbalance by analyzing the local data distribution. As MFE Loss ($L_{M\!F\!E}$) is based on \textbf{Mean-Square-Error Loss} ($L_{M\!S\!E}$), we regard $L_{M\!S\!E}$ as the baseline. As stated in the introduction, applying $L_{M\!F\!E}$ requires knowing what minority classes specifically are. In FL, such information is difficult to acquire globally, and the standard method based on $L_{M\!F\!E}$ can only analyze the local data of each client. 

\begin{figure}[h!]
    \centering
    \includegraphics[width=.75\linewidth]{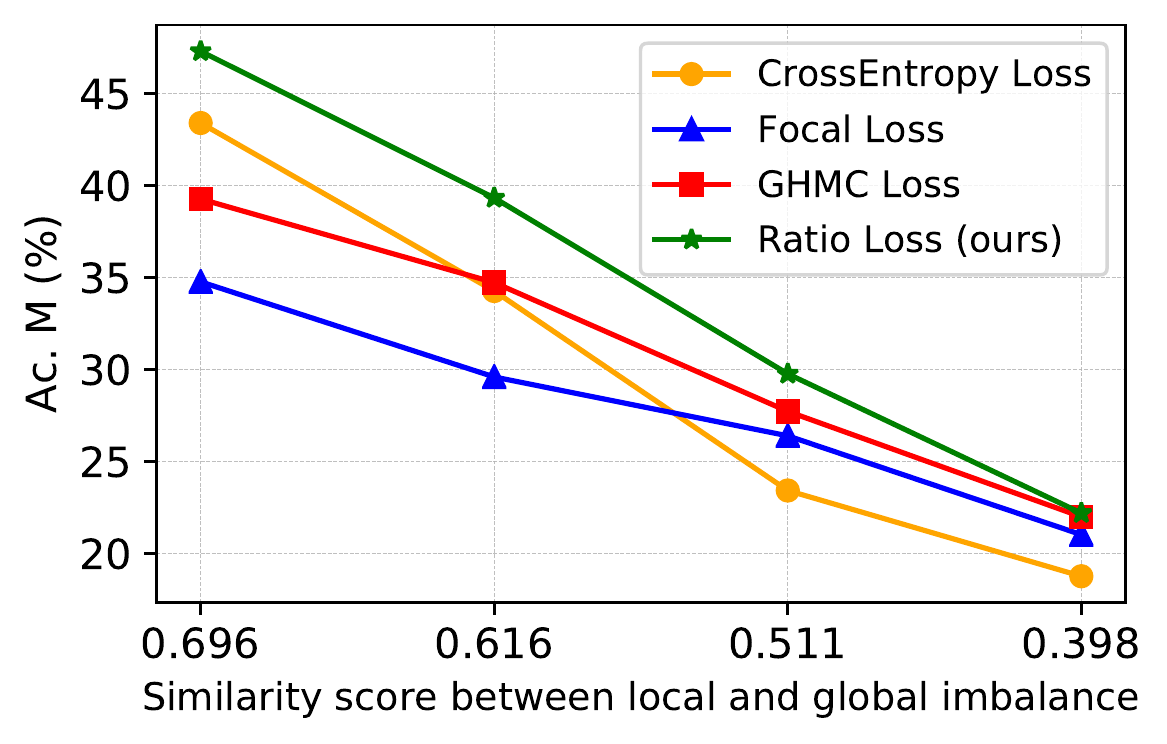}
    \caption{Comparison between Ratio Loss, CrossEntropy Loss, Focal Loss and GHMC Loss under different levels of mismatch between local and global imbalance.}
    \label{Res_mismatch}
\end{figure}
Table~\ref{Sup_MFE} shows the comparison between $L_{M\!S\!E}$ and $L_{M\!F\!E}$. The global imbalance ratio is $\Gamma\!=\!10\!:\!1$ (Ac.M degrades to zero when the ratio is larger than $10\!:\!1$). From the results, we can clearly see that using $L_{M\!F\!E}$ locally has similar performance as the baseline. Moreover, we can observe that the performance of global model with $L_{M\!F\!E}$ is worse than the baseline when the cosine similarity (CS) score is very low. This indicates the negative impact of solving the imbalance locally to the global model when there is significant mismatch between the local and global imbalance. From these results, we can see the necessity of estimating the global imbalance from the perspective of the model rather than the data distribution, which is consistent with the principles of Focal Loss, GHMC Loss and Ratio Loss.

Fig.~\ref{Res_mismatch} further shows the Ac.M of four methods under different levels of mismatch for CIFAR10 with $\Gamma\!=\!10\!:\!1$ (more results for other datasets and global imbalance are in the $\mathcal{S\!M}$). The x-axis shows the average mismatch between local and global imbalance, measured by the average of CS scores over clients (the larger the number, the less the mismatch). From the figure, we can observe that 1) larger mismatch worsens the performance for all methods, and 2) our method outperforms the other methods under all levels of mismatch.

%% file: Conclusion.tex
\section{Conclusion}
We present a novel method to address the class imbalance issue in federate learning (FL). Our approach includes a monitoring scheme that can infer the composition of training data at every FL round and detect the existence of possible global imbalance, and a new loss function (Ratio Loss) for mitigating the impact of global class imbalance. Extensive experiments demonstrate that our method can significantly outperform previous imbalance solutions in its accuracy of classifying minority classes and meanwhile not impact the performance of majority classes in FL. Even in the setting of regular neural network training, our method can also achieve the state-of-art performance. And considering the security concern in FL, our method works effectively without the sacrifice of user privacy.

%% file: Acknowledgment.tex
\section{Acknowledgment}
We appreciate all anonymous reviewers for their valuable and detailed comments. We gratefully acknowledge the support from NSF grants 1834701, 1839511 and 1724341.

%% file: Supplementary.tex
\appendix
\section{Supplementary Material}
\subsection{Foreword}

This section contains additional details for the submitted article \textit{Addressing Class Imbalance in Federated Learning}, including mathematical proofs, experimental details and further discussions. The implementation code can be found in \url{https://github.com/balanced-fl/Make_FL_More_Balanced}.

\subsection{Proof of Theorem 1}
\smallskip
\noindent\textbf{Theorem 1}: \emph{For any real-valued neural network $f$ whose last layer is a linear layer with a softmax operation (without any parameter sharing), and for any input sample $X_i^{(p)}$ and $X_j^{(p)}$ of the same class $p$, if the inputs of the last layer $Y_i$ and $Y_j$ are identical, the gradients of link weights $\mathcal{W}$ between the last layer and its former layer induced by $X_i^{(p)}$ and $X_j^{(p)}$ are identical.}
\begin{equation}
    Y_i = Y_j \Rightarrow \nabla_{\mathcal{W}}^i L (\mathbb{W}) = \nabla_{\mathcal{W}}^j L (\mathbb{W})
\end{equation}

\smallskip
\noindent\textbf{Proof}: If the last layer is linear, the output before the softmax operation is of the form $z_{(i)} = \mathcal{W}_{(i)} \cdot Y + b$, where $Y$ is the output of the former layer and $b$ is the bias of the last layer. The softmax operation can be written as $f_{(i)}=Softmax(z_{(i)})=e^{z_{(i)}} / \left (\sum_{i=1}^{Q} e^{z_{(i)}}\right )$. In addition, samples $\mathcal{X}_i$ and $\mathcal{X}_j$ belong to class $p$, and thus their corresponding loss is $L(\mathbb{W}) = L_{C\!E}(p) = - log(f_{(p)})$. We know that $\mathcal{W}$ is a matrix with the size $Q \times s$, where $s$ is the number of neurons for the former layer. For every component $\mathcal{W}_{(m, n)}$ (where $m\!=\!1,...,Q$ and $n\!=\!1,...,s$) in $\mathcal{W}$, following the principle of back-propagation, we have 
\begin{equation}
\begin{split}
    \nabla_{\mathcal{W}_{(m,n)}} L (\mathbb{W}) &= \frac{\partial L(\mathbb{W})}{\partial f_{(p)}} \cdot \frac{\partial f_{(p)}}{\partial z_{(m)}} \cdot \frac{\partial z_{(m)}}{\partial \mathcal{W}_{(m,n)}}\\
    &= -\frac{1}{f_{(p)}} \cdot \frac{\partial f_{(p)}}{\partial z_{(m)}} \cdot y_{(n)}
\end{split}
\end{equation}
where
\begin{equation}
\begin{split}
    \frac{\partial f_{(p)}}{\partial z_{(m)}} = \frac{\partial}{\partial z_{(m)}}\left (\frac{e^{z_{(p)}}}{\sum_{i=1}^{Q}e^{z_{(i)}}}\right )
\end{split}
\end{equation}
Consider two cases, if $p=m$: 
\begin{equation}
\begin{split}
    \frac{\partial f_{(p)}}{\partial z_{(p)}} &= \frac{e^{z_{(p)}}}{\sum_{i=1}^{Q}e^{z_{(i)}}} - \frac{e^{z_{(p)}}}{\sum_{i=1}^{Q}e^{z_{(i)}}} \cdot \frac{e^{z_{(p)}}}{\sum_{i=1}^{Q}e^{z_{(i)}}}\\
    &= f_{(p)}(1-f_{(p)})
\end{split}
\end{equation}
if $p \neq m$: 
\begin{equation}
\begin{split}
    \frac{\partial f_{(p)}}{\partial z_{(m)}} &=  -  \frac{e^{z_{(p)}}}{\sum_{i=1}^{Q}e^{z_{(i)}}} \cdot \frac{e^{z_{(m)}}}{\sum_{i=1}^{Q}e^{z_{(i)}}}\\
    &= -f_{(p)} \cdot f_{(m)}
\end{split}
\end{equation}

Therefore, when $p=m$, $\nabla_{\mathcal{W}_{(m,n)}} L (\mathbb{W}) = (f_{(p)} - 1) y_{(n)}$. If $p \neq m$, $\nabla_{\mathcal{W}_{(m,n)}} L (\mathbb{W}) = - f_{(m)} y_{(n)}$. Here, $f_{(p)} = \mathcal{W}_{(p)} \cdot Y + b$, and $f_{(m)} = \mathcal{W}_{(m)} \cdot Y + b$. For the given neural network, parameters ($\mathcal{W}_{i}, (i=1,...,Q)$ and $b$) of the last layer are fixed. For different samples $\mathcal{X}_i^{(p)}$ and $\mathcal{X}_j^{(p)}$, if their corresponding outputs $Y_i$ and $Y_j$ are identical, then $f_{(p)}^i$ and $f_{(p)}^j$ as well as $f_{(m)}^i$ and $f_{(m)}^j$ are also identical. Thus, each component in $\mathcal{W}$ holds true for $\nabla_{\mathcal{W}_{(m,n)}}^i L (\mathbb{W}) = \nabla_{\mathcal{W}_{(m,n)}}^j L (\mathbb{W})$, and $\nabla_{\mathcal{W}}^i L (\mathbb{W}) = \nabla_{\mathcal{W}}^j L (\mathbb{W})$. $\hfill \blacksquare$

\subsection{Proof of Theorem 2}

\smallskip
\noindent\textbf{Theorem 2}: \emph{For any real-valued neural network $f$ whose last layer is a linear layer with a softmax operation (without any parameter sharing), and the activation function between the last layer and its former layer is non-negative (e.g., Sigmoid and ReLu), if $f$ has learned imbalanced dataset, for any majority class $\mathcal{A}$, any minority class $\mathcal{B}$, and another class $\mathcal{C}$ ($\mathcal{C} \neq \mathcal{A} \,\text{and}\, \mathcal{C} \neq \mathcal{B}$, but $\mathcal{C}$ can be any class other than chosen $\mathcal{A}\, \&\, \mathcal{B}$) fed to $f$, we have:}
\begin{equation}
    |\nabla_{\mathcal{W}_{(\mathcal{A})}^{(\mathcal{C})}} L(\mathbb{W})| > |\nabla_{\mathcal{W}_{(\mathcal{B})}^{(\mathcal{C})}} L(\mathbb{W})|
\end{equation}

\smallskip
\noindent\textbf{Assumption}: 1) The input similarity between class $\mathcal{C}$ and $\mathcal{A}$ is the same as between class $\mathcal{C}$ and $\mathcal{B}$. 2) The reason why there is classification accuracy degradation on the minority class $\mathcal{B}$ is that its probability result $f_{(\mathcal{B})}^{(\mathcal{B})}$ is not distinguishable with the output of other $f_{(i)}^{(\mathcal{B})}$ ($i=1,...,Q\, \mbox{and} \, i \neq \mathcal{B}$). Thus minority classes can be regarded as hard samples generally, while majority classes are easy samples, i.e.,
\begin{equation}
    \frac{f_{(\mathcal{A})}^{(\mathcal{A})}}{f_{(\mathcal{B})}^{(\mathcal{A})}} \gg \frac{f_{(\mathcal{B})}^{(\mathcal{B})}}{f_{(\mathcal{A})}^{(\mathcal{B})}} > 1
    \label{theorem2_relation}
\end{equation}

\smallskip
\noindent\textbf{Proof}: Under the same circumstances as Theorem 1, we can formulate the vector form of $\nabla_{\mathcal{W}_{(\mathcal{A})}^{(\mathcal{C})}} L(\mathbb{W})$ as $\nabla_{\mathcal{W}_{(\mathcal{A})}^{(\mathcal{C})}} L(\mathbb{W})=\left[\nabla_{\mathcal{W}_{(\mathcal{A}, 1)}^{(\mathcal{C})}} L(\mathbb{W}),...,\nabla_{\mathcal{W}_{(\mathcal{A}, s)}^{(\mathcal{C})}} L(\mathbb{W}) \right ]$. Because $\mathcal{C} \neq \mathcal{A}$, we have each component in $\nabla_{\mathcal{W}_{(\mathcal{A})}^{(\mathcal{C})}} L(\mathbb{W})$:
\begin{equation}
    \nabla_{\mathcal{W}_{(\mathcal{A}, 1)}^{(\mathcal{C})}} L(\mathbb{W}) = - f_{(\mathcal{A})}^{(\mathcal{C})} y_{(i)}
    \label{theorem2_gradient}
\end{equation}
where $i=1, ..., s$. For $\nabla_{\mathcal{W}_{(\mathcal{B})}^{(\mathcal{C})}} L(\mathbb{W})$, the equation is similar with the only difference on the subscription. When comparing Eq.~(\ref{theorem2_gradient}) of classes $\mathcal{A}$ and $\mathcal{B}$, since $y_i$ is identical, we only need to compare $f_{(\mathcal{A})}^{(\mathcal{C})}$ and $f_{(\mathcal{B})}^{(\mathcal{C})}$ when the input is  samples of class $\mathcal{C}$. 
Follow the definition of taking the gradient kernel to evaluate the similarity between two inputting samples~\cite{charpiat2019input}, the Assumption 1) can be formulated as:
\begin{equation}
    Y^{(\mathcal{C})} \cdot Y^{(\mathcal{A})} = Y^{(\mathcal{C})} \cdot Y^{(\mathcal{B})}
    \label{theorem2_equal}
\end{equation}
And for assumption 2), with the softmax operation, we can obtain the following from Eq.~(\ref{theorem2_relation}):
\begin{equation}
    \frac{f_{(\mathcal{A})}^{(\mathcal{A})}}{f_{(\mathcal{B})}^{(\mathcal{A})}} = \frac{e^{z_{(\mathcal{A})}^{(\mathcal{A})}}}{e^{z_{(\mathcal{B})}^{(\mathcal{A})}}} > 1 \Rightarrow Y^{(\mathcal{A})} \cdot \mathcal{W}_{(\mathcal{A})} - Y^{(\mathcal{A})} \cdot \mathcal{W}_{(\mathcal{B})} > 0
    \label{theorem2_vector_A}
\end{equation}
\begin{equation}
    \frac{f_{(\mathcal{B})}^{(\mathcal{B})}}{f_{(\mathcal{A})}^{(\mathcal{B})}} = \frac{e^{z_{(\mathcal{B})}^{(\mathcal{B})}}}{e^{z_{(\mathcal{A})}^{(\mathcal{B})}}} > 1 \Rightarrow Y^{(\mathcal{B})} \cdot \mathcal{W}_{(\mathcal{B})} - Y^{(\mathcal{B})} \cdot \mathcal{W}_{(\mathcal{A})} > 0
    \label{theorem2_vector_B}
\end{equation}
Based on the relation shown in Eq.~(\ref{theorem2_relation}), we consider Eq.~(\ref{theorem2_vector_A}) minus Eq.~(\ref{theorem2_vector_B}), and get the following result:
\begin{equation}
    \left(Y^{(\mathcal{A})} + Y^{(\mathcal{B})}\right) \cdot (\mathcal{W}_{(\mathcal{A})} - \mathcal{W}_{(\mathcal{B})}) > 0
    \label{theorem2_non-negative}
\end{equation}
After formulating these two assumptions properly, we divide $f_{(\mathcal{A})}^{(\mathcal{C})}$ by $f_{(\mathcal{B})}^{(\mathcal{C})}$ as:
\begin{equation}
    \frac{f_{(\mathcal{A})}^{(\mathcal{C})}}{f_{(\mathcal{B})}^{(\mathcal{C})}} = e^{z_{(\mathcal{A})}^{(\mathcal{C})} - z_{(\mathcal{B})}^{(\mathcal{C})}} = e^{Y^{(\mathcal{C})} \cdot (\mathcal{W}_{(A)} - \mathcal{W}_{(B)})}
\end{equation}
To prove $f_{(\mathcal{A})}^{(\mathcal{C})} > f_{(\mathcal{B})}^{(\mathcal{C})}$, we just need to demonstrate that $Y^{(\mathcal{C})} \cdot (\mathcal{W}_{(A)} - \mathcal{W}_{(B)})$ is positive, as shown below. 

First, since the activation function between the last layer and its former layer is non-negative, every component of $Y$ is non-negative, and thus Eq.~(\ref{theorem2_equal}) is positive. If we multiply $Y^{(\mathcal{C})} \cdot (\mathcal{W}_{(A)} - \mathcal{W}_{(B)})$ with the left side of Eq.~(\ref{theorem2_non-negative}), we have:
\begin{equation}
    Y^{(\mathcal{C})} \cdot (\mathcal{W}_{(A)} \!-\! \mathcal{W}_{(B)}) \cdot \left(Y^{(\mathcal{A})} \!+\! Y^{(\mathcal{B})}\right) \cdot (\mathcal{W}_{(\mathcal{A})} \!-\! \mathcal{W}_{(\mathcal{B})})
    \label{theorem2_final_vector}
\end{equation}
As every part in Eq.~(\ref{theorem2_final_vector}) is a vector with the same size of $s$ (the neuron number of the former layer), we can change the multiplication sequence and obtain a new form as:
\begin{equation}
\begin{split}
    &Y^{(\mathcal{C})} \cdot \left(Y^{(\mathcal{A})} \!+\! Y^{(\mathcal{B})} \right) \cdot \left(\mathcal{W}_{(\mathcal{A})} \!-\! \mathcal{W}_{(\mathcal{B})} \right)^2 \\
    &= \left(Y^{(\mathcal{C})} \cdot Y^{(\mathcal{A})} + Y^{(\mathcal{C})} \cdot Y^{(\mathcal{B})}\right) \cdot \left(\mathcal{W}_{(\mathcal{A})} \!-\! \mathcal{W}_{(\mathcal{B})} \right)^2
    \label{theorem2_final_result}
\end{split}
\end{equation}
The two parts in Eq.~(\ref{theorem2_final_result}) are both positive, and thus Eq.~(\ref{theorem2_final_result}) is positive and greater than zero. Then, since the left side of Eq.~\eqref{theorem2_non-negative} is positive, we can conclude that $Y^{(\mathcal{C})} \cdot (\mathcal{W}_{(A)} - \mathcal{W}_{(B)})$ is positive, which proves that $f_{(\mathcal{A})}^{(\mathcal{C})} > f_{(\mathcal{B})}^{(\mathcal{C})}$.
$\hfill \blacksquare$

\subsection{Empirical Study}
\begin{figure}[htb]
    \centering
    \includegraphics[width=1.\linewidth]{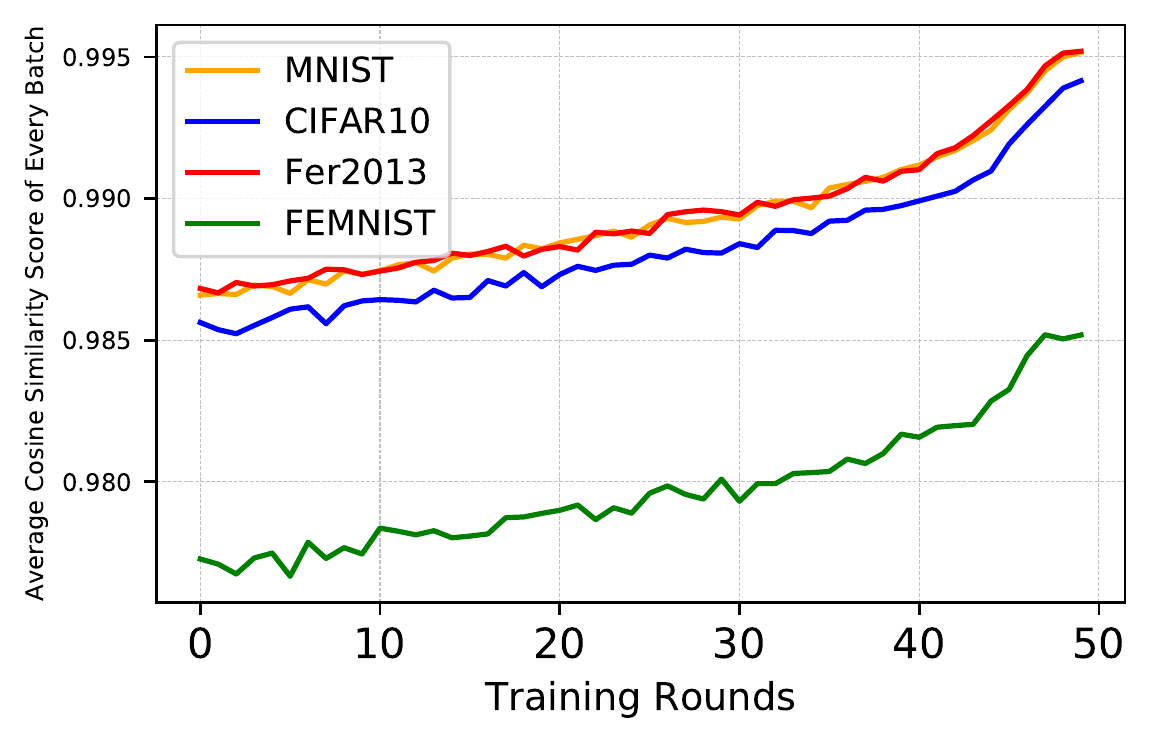}
    \caption{Average cosine similarity score of all pairs of $Y$s for the same class within every batch in a global round during FL training, over four datasets.}
    \label{Sup_empirical_sim}
\end{figure}

\begin{figure}[htb]
    \centering
    \includegraphics[width=1.\linewidth]{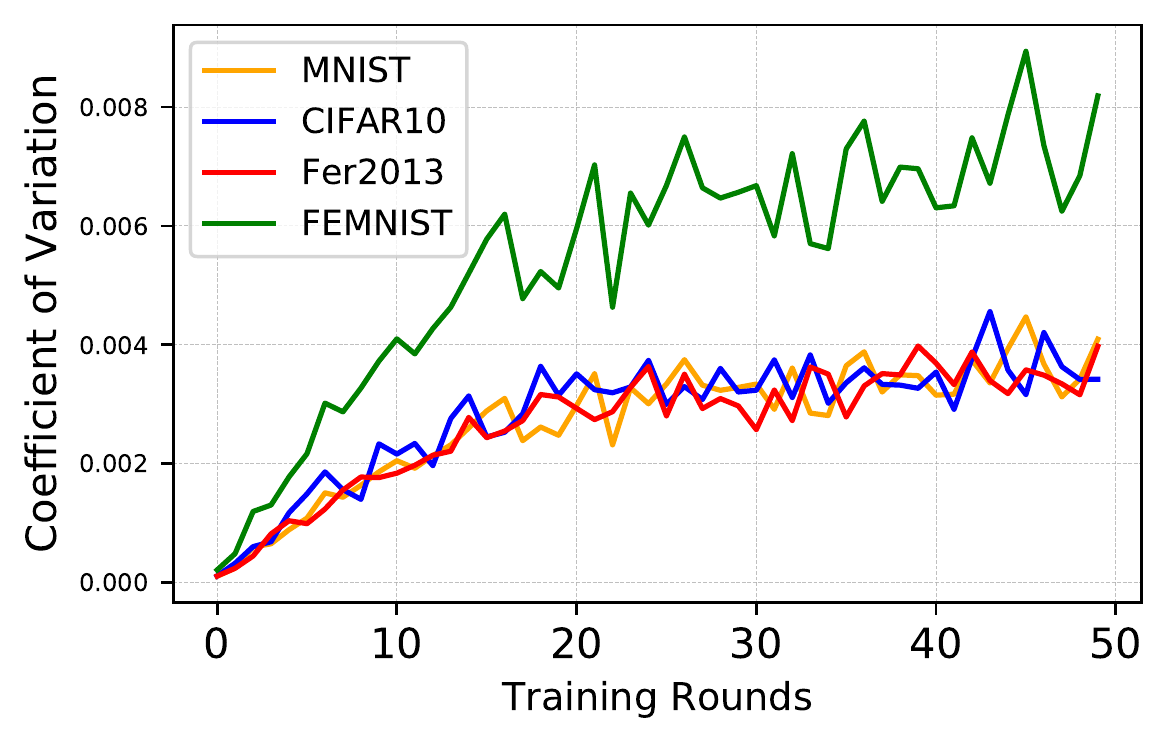}
    \caption{Average coefficient of variation for multiplications between all pairs of $Y$ for the same class within every batch in a global round during FL training, over four datasets.}
    \label{Sup_empirical_mul}
\end{figure}

In the regular supervised learning, there is great possibility that data samples of a same class will induce very similar classification results. Here, we conduct an empirical study to extend such similarities to the inputs of the last linear layer. For all four datasets we use in our evaluation experiments, we set an additional client in the FL training process, and this client will feed the overall training data to compare the inputs of the last layer $Y$. We get the $Y$ of every batch at every epoch, compute the cosine similarity ($C\!S$) between every pair of $Y_i, Y_j$ within a same batch, and multiply them to obtain the multiplication of vectors ($M$). We think that $C\!S$ can compare two vectors in terms of Vector Directivity, while $M$ can compare the similarity of Vector Distance. Fig.~\ref{Sup_empirical_sim} presents the cosine similarity score during FL training. We can clearly see that the cosine similarity score is very high and close to 1, which demonstrates that $Y$ of different samples for the same class is very similar in terms of Vector Directivity. As the training continues, we can observe that the similarity score is rising. Fig.~\ref{Sup_empirical_mul} shows the coefficient of variation (the ratio between the standard deviation and the average value) for $M$ during training, and this metric can properly describe the dispersion degree of $M$ from its mean value. As we can see from the figure, the coefficient of variation is extremely low across the whole training process, which implies that nearly all $M$s concentrate near its average value. This finding strongly supports the design of our monitoring scheme.

\begin{table*}[thb]
\centering
% \small
\begin{tabular}{c|ccccccccccc}
\hline
$T_{Ra}$                    & 1.00  & 1.05  & 1.10  & 1.15  & 1.20  & \textbf{1.25}  & 1.30  & 1.40  & 1.50  & 2.00  & 5.00  \\ \hline
\textbf{Mean}        & .9877 & .9860 & .9890 & .9893 & .9897 & \textbf{.9899} & .9885 & .9879 & .9882 & .9877 & .9870 \\ \hline
\textbf{Var. ($\times\!10^{-5}$)} & 4.340 & 6.190 & 3.260 & 2.944 & 2.697 & \textbf{1.345} & 4.218 & 3.613 & 4.183 & 8.099 & 28.86 \\ \hline
\end{tabular}
\caption{The mean and variance of the similarity scores between our estimation of the data composition and the ground truth over MNIST under different $T_{Ra}$.}
\label{Sup_hyper_T_ra}
\end{table*}

\begin{table*}[htb]
\centering
\begin{tabular}{c|cccccccccc}
\hline
$\alpha$                                                          & 0.25   & 0.50   & 0.75   & \textbf{1.00}   & 1.25   & 1.50   & 1.75   & 2.00   & 3.00   & 5.00   \\ \hline
\textbf{\begin{tabular}[c]{@{}c@{}}AC.M\\ \%\end{tabular}} & 77.31  & 80.68  & 78.56  & \textbf{82.52}  & 81.82  & 81.06  & 75.82  & 80.64  & 75.05  & 41.12  \\ \hline
\textbf{AUC}                                               & .9579 & .9588 & .9608 & \textbf{.9692} & .9663 & .9650 & .9562 & .9642 & .9519 & .8906 \\ \hline
\end{tabular}
\caption{Performance comparison under different $\alpha$ values ($\beta=0.10$, global imbalance $\Gamma=50:1$).}
\label{Sup_hyper_a}
\end{table*}

\begin{table*}[h!]
\centering
\begin{tabular}{c|ccccccccc}
\hline
$\beta$                                                          & 0.02   & 0.04   & 0.06   & 0.08   & \textbf{0.10}   & 0.12   & 0.15   & 0.20   & 0.30   \\ \hline
\textbf{\begin{tabular}[c]{@{}c@{}}AC.M\\ \%\end{tabular}} & 80.44  & 76.84  & 78.66  & 79.41  & \textbf{82.52}  & 82.34  & 80.85  & 79.65  & 74.40  \\ \hline
\textbf{AUC}                                               & .9641 & .9575 & .9608 & .9626 & \textbf{.9692} & .9675 & .9645 & .9604 & .9496 \\ \hline
\end{tabular}
\caption{Performance comparison under different $\beta$ values ($\alpha=1.00$, global imbalance $\Gamma=50:1$).}
\label{Sup_hyper_b}
\end{table*}

\subsection{More Experiment Results}
\smallskip
\noindent \textbf{$T_{Ra}$ of Monitoring Scheme}

As explained in the main paper, to build our filter that selects weights whose updating magnitudes are relatively large, we set a threshold $T_{Ra}$, with $T_{Ra} \!=\! 1.25$ in our experiments. The value of $T_{Ra}$ is chosen based on a set of calibration experiments, where the dataset is MNIST and other settings are the same with those of the main paper. The results of these experiments are presented in Table~\ref{Sup_hyper_T_ra}, showing the mean and variance of cosine similarities for different values of $T_{Ra}$. 
We can see that $T_{Ra} \!=\! 1.25$ provides the largest mean and the smallest variance.

\smallskip
\noindent \textbf{Hyper-parameters of Ratio Loss}

As stated in the paper, the hyper-parameters $\alpha$ and $\beta$ in the Ratio Loss function are set to $1.0$ and $0.1$,  respectively. This is based on our calibration experiments on MNIST, as shown in Tables~\ref{Sup_hyper_a} and~\ref{Sup_hyper_b}. The calibration experiments have the same setting as in the paper, with the global imbalance ratio set as $\Gamma=50:1$.

\smallskip
\noindent \textbf{The Importance of Early Acknowledgment}

We would like to demonstrate the importance of acknowledging the class imbalance as early as possible. Table~\ref{Importance_early} shows the different results when the global class imbalance ($\Gamma=100\!:\!1$) is acknowledged at the start ($10$-th epoch out of $50$ epochs, denoted as S.), middle ($30$-th epoch, M.), or towards the end ($45$-th epoch, E.) of FL training. We assume that once the global imbalance is acknowledged, the administrator ($\mathcal{AD}$) will replace the imbalanced data set with a balanced one. We can see that earlier acknowledgment can greatly help improve the performance.

\begin{table}[t]
\centering
\small
\setlength{\tabcolsep}{1.5mm}{
\begin{tabular}{c|c|cccc}
\hline
\textbf{Metric}                                                             & \textbf{Stage} & \textbf{MNIST} & \textbf{CIFAR10} & \textbf{Fer2013} & \textbf{FEMNIST} \\ \hline
\multirow{3}{*}{\textbf{\begin{tabular}[c]{@{}c@{}}Ac.M\\ \%\end{tabular}}} & E.             & 31.64          & 03.28            & 06.99            & 22.37            \\
                                                                            & M.             & 97.01          & 57.73            & 92.59            & 90.18            \\
                                                                            & S.             & \textbf{97.81} & \textbf{68.27}   & \textbf{99.67}   & \textbf{96.23}   \\ \hline
\multirow{3}{*}{\textbf{AUC}}                                               & E.             & .8816          & .6893            & .7648            & .8424            \\
                                                                            & M.             & .9896          & .7436            & .9463            & .9653            \\
                                                                            & S.             & \textbf{.9903} & \textbf{.7706}   & \textbf{.9970}   & \textbf{.9850}   \\ \hline
\end{tabular}
}
\caption{Comparison when global imbalance is acknowledged at different stages of FL training.}
\label{Importance_early}
\end{table}

\smallskip
\noindent \textbf{Local Batch Size and Epochs}

To investigate the impact of different local training hyper-settings (local batch size and local epochs), we evaluate our monitoring scheme under various situations. Table~\ref{Sup_bs_ep} presents the mean and variance of cosine similarities between our estimation of data composition and the ground truth with different local batch sizes ($32, 64, 96, 128, 256$) over CIFAR10 and FEMNIST, respectively. The results of different local epochs can also be found in Table~\ref{Sup_bs_ep}. According to these results, the cosine similarity still remains extremely high in varying cases of different local batch sizes and epochs. In addition, the relatively small variance demonstrates that the performance of our monitoring scheme is stable without great fluctuation.

\begin{table*}[htb]
\centering
% \small
\begin{tabular}{c|ccccc|ccccc}
\hline
\textbf{Data}         & \multicolumn{5}{c|}{\textbf{CIFAR10}} & \multicolumn{5}{c}{\textbf{FEMNIST}}  \\ \hline
\textbf{Batch Size}   & 32    & 64    & 96    & 128   & 256   & 32    & 64    & 96    & 128   & 256   \\ \hline
\textbf{Mean}         & .9797 & .9815 & .9831 & .9838 & .9884 & .9802 & .9824 & .9785 & .9804 & .9790 \\
\textbf{Var. $\times \! 10^{-5}$}         & 14.34 & 7.837 & 13.72 & 4.581 & 5.044 & 2.924 & 5.008 & 3.093 & 3.687 & 2.619 \\ \hline
\textbf{Local Epochs} & 4     & 8     & 10    & 20    & 30    & 4     & 8     & 10    & 20    & 30    \\ \hline
\textbf{Mean}         & .9828 & .9803 & .9789 & .9797 & .9799 & .9794 & .9807 & .9828 & .9824 & .9869 \\
\textbf{Var. $\times \! 10^{-5}$}         & 4.457 & 8.555 & 18.68 & 9.150 & 7.096 & 5.276 & 3.460 & 2.963 & 3.867 & 2.454 \\ \hline
\end{tabular}
\caption{The mean and variance of the similarity scores between our estimation of the data composition and the ground truth over CIFAR and FEMNIST, respectively, under different local batch sizes and epochs.}
\label{Sup_bs_ep}
\end{table*}

\smallskip
\noindent \textbf{Different Number of Selected Clients}

\begin{figure}[h]
    \centering
    \includegraphics[width=0.8\linewidth]{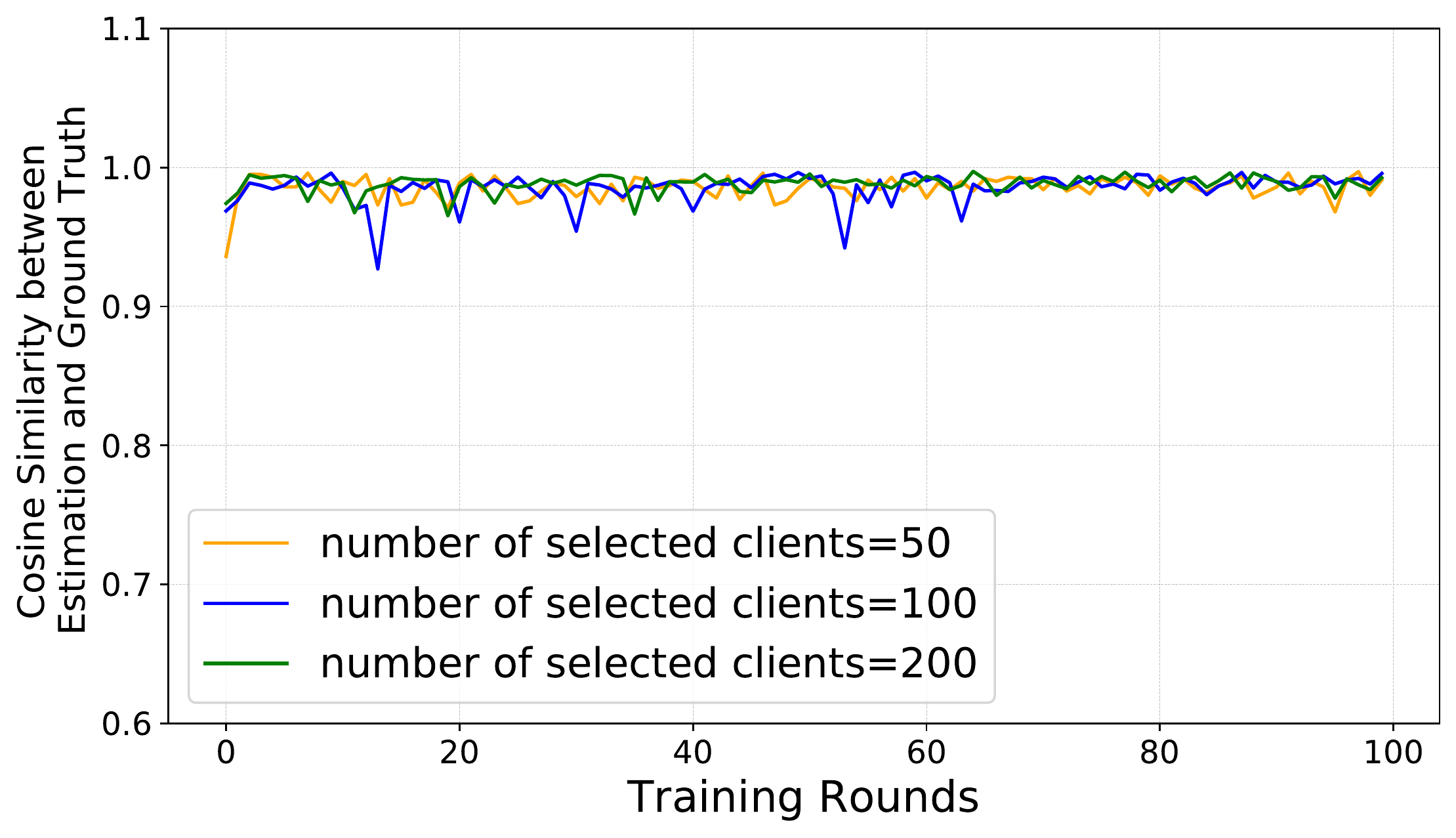}
    \caption{Similarity score between our estimation of the data composition and the ground truth over CIFAR10, under different number of selected clients at each round.}
    \label{Sup_client_num_CIFAR10}
\end{figure}

\begin{figure}[h]
    \centering
    \includegraphics[width=0.8\linewidth]{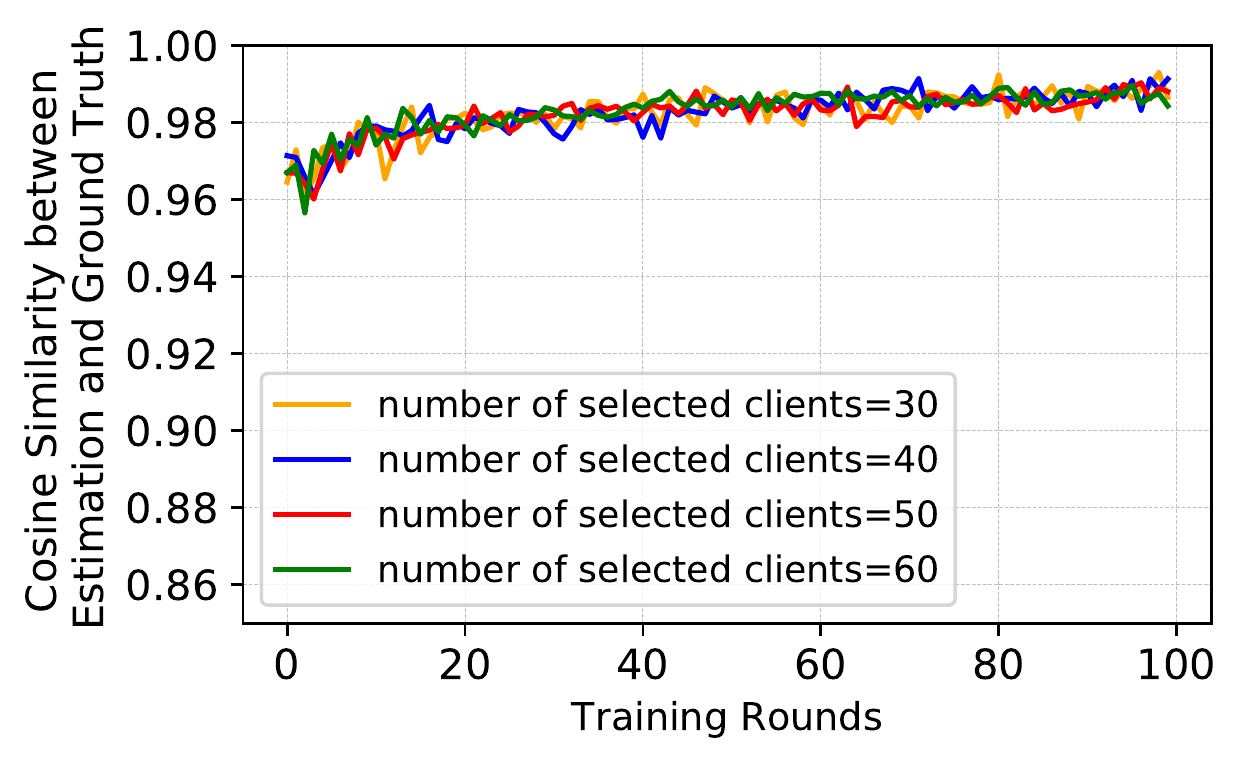}
    \caption{Similarity score between our estimation of the data composition and the ground truth over FEMNIST, under different number of selected clients at each round.}
    \label{Sup_client_num_FEMNIST}
\end{figure}

To further evaluate the effectiveness of our monitoring scheme, we carry out experiments with different number of participating clients at each ground on CIFAR10 and FEMNIST. For CIFAR10, we set the overall number of clients to $1000$, and randomly select $50, 100, 200$ of them at each round, respectively, to upload their gradient updates. The results are shown in Fig.~\ref{Sup_client_num_CIFAR10}. We also randomly select $30, 40, 50, 60, 80$ clients, respectively, to aggregate their gradient updates for FEMNIST. The results are presented in Fig.~\ref{Sup_client_num_FEMNIST}. We can see that with the increase of the selected client number, the similarity score between our estimation of the data composition and the ground truth remains very high and even becomes closer to 1, which helps further improve the performance of our approach. 

\smallskip
\noindent \textbf{Classification Accuracy of Majority Classes}

Here, we present the average classification accuracy of majority classes for the experiments of comparison among methods based on Ratio Loss, CrossEntropy Loss, Focal Loss, and GHMC Loss. Please refer to Table~\ref{Sup_majority} and~\ref{Sup_majority_femnist} for the results. We can observe that our Ratio Loss will not sacrifice the performance of majority classes to mitigate the negative impact of class imbalance, which also proves the effectiveness of our method.

\begin{table*}[htb]
\centering
\small
\setlength{\tabcolsep}{1.45mm}{
\begin{tabular}{c|c|ccccc|ccccc|ccccc}
\hline
\multicolumn{2}{c|}{\textbf{Data}}                                             & \multicolumn{5}{c|}{\textbf{MNIST}}   & \multicolumn{5}{c|}{\textbf{CIFAR10}} & \multicolumn{5}{c}{\textbf{Fer2013}}  \\ \hline
\multicolumn{2}{c|}{$\Gamma$}                                            & B.    & 10:1  & 20:1  & 50:1  & 100:1 & B.    & 10:1  & 20:1  & 50:1  & 100:1 & B.    & 10:1  & 20:1  & 50:1  & 100:1 \\ \hline
\multirow{4}{*}{\textbf{\begin{tabular}[c]{@{}c@{}}Ac.\\ \%\end{tabular}}} & $L_{C\!E}$ & 98.28 & 97.69 & 97.58 & 97.55 & 97.23 & 63.03 & 61.22 & 61.02 & 58.90 & 58.61 & 97.99 & 96.28 & 96.21 & 95.98 & 94.99 \\
                                                                           & $L_{F\!L}$ & 97.03 & 97.28 & 97.14 & 97.10 & 97.07 & 53.34 & 52.78 & 52.43 & 52.18 & 52.10 & 89.37 & 88.24 & 88.17 & 86.89 & 85.87 \\
                                                                           & $L_{G\!H\!M\!C}$ & 96.01 & 95.81 & 95.63 & 95.37 & 95.28 & 51.72 & 51.99 & 51.98 & 50.60 & 50.19 & 76.27 & 75.49 & 75.20 & 74.48 & 74.23 \\
                                                                           & $L_{R\!L}$(ours) & 98.11 & 97.99 & 97.94 & 97.55 & 97.13 & 63.03 & 62.47 & 62.34 & 58.80 & 57.99 & 95.66 & 93.23 & 92.98 & 92.13 & 91.70 \\ \hline
\end{tabular}
}
\caption{Comparison of the classification accuracy on the majority classes between our method ($L_{R\!L}$) and previous methods based on CrossEntropy Loss ($L_{C\!E}$), Focal Loss ($L_{F\!L}$) and GHMC Loss ($L_{G\!H\!M\!C}$) in federate learning, over three datasets and different levels of global imbalance.}
\label{Sup_majority}
\end{table*}

\begin{table}[H]
\centering
\small
\setlength{\tabcolsep}{1.9mm}{
\begin{tabular}{c|c|ccccc}
\hline
\multicolumn{2}{c|}{\textbf{Data}}                                             & \multicolumn{5}{c}{\textbf{FEMNIST}}  \\ \hline
\multicolumn{2}{c|}{$\Gamma$}                                               & B.    & 10:1  & 20:1  & 50:1  & 100:1 \\ \hline
\multirow{4}{*}{\textbf{\begin{tabular}[c]{@{}c@{}}Ac.\\ \%\end{tabular}}} & $L_{C\!E}$ & 91.28 & 90.94 & 90.28 & 90.17 & 90.08 \\
                                                                           & $L_{F\!L}$ & 92.23 & 91.79 & 91.66 & 91.10 & 90.97 \\
                                                                           & $L_{G\!H\!M\!C}$ & 93.88 & 93.29 & 93.01 & 92.83 & 92.77 \\
                                                                           & $L_{R\!L}$(ours) & 93.97 & 93.26 & 93.18 & 93.07 & 92.93 \\ \hline
\end{tabular}
}
\caption{Comparison of the classification accuracy on the majority classes between our method ($L_{R\!L}$) and previous methods based on CrossEntropy Loss ($L_{C\!E}$), Focal Loss ($L_{F\!L}$) and GHMC Loss ($L_{G\!H\!M\!C}$) in federate learning, over FEMNIST and different levels of global imbalance.}
\label{Sup_majority_femnist}
\end{table}

\smallskip
\noindent \textbf{Convergence Curve}

\begin{figure}[htb]
    \centering
    \includegraphics[width=0.9\linewidth]{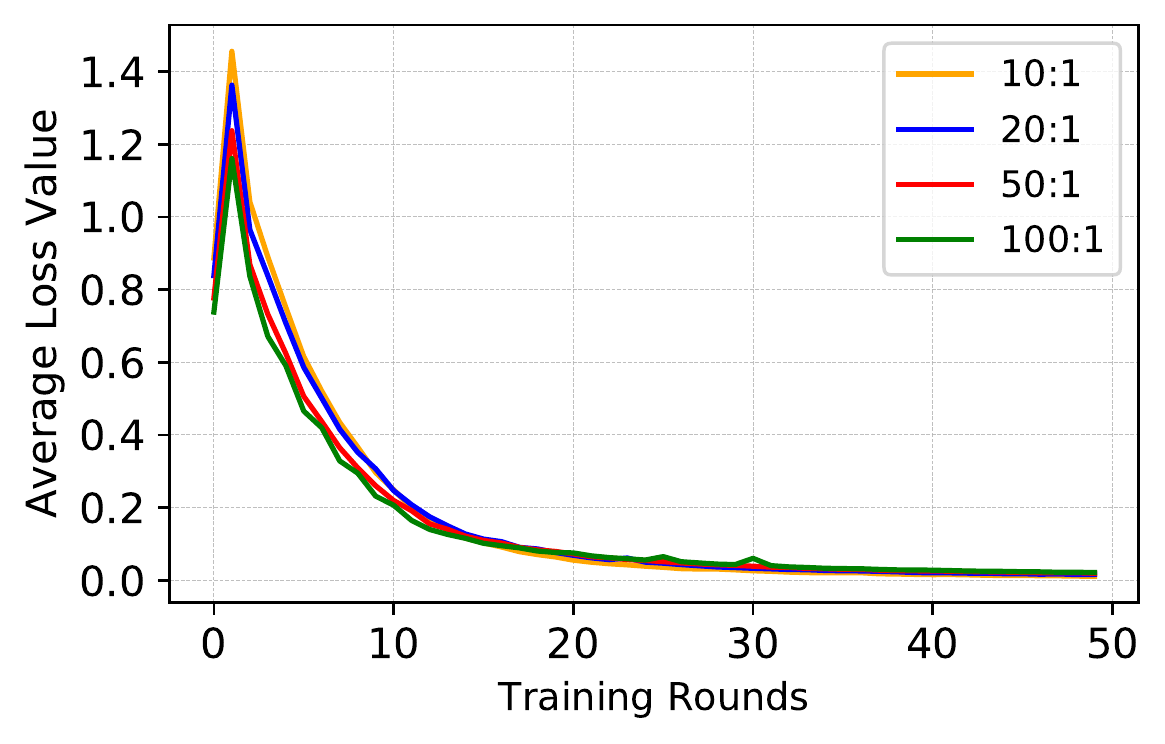}
    \caption{Average loss value of Ratio Loss with different global imbalance ratios during FL training over CIFAR10.}
    \label{Sup_curve_cifar_ratio}
\end{figure}

\begin{figure}[htb]
    \centering
    \includegraphics[width=0.9\linewidth]{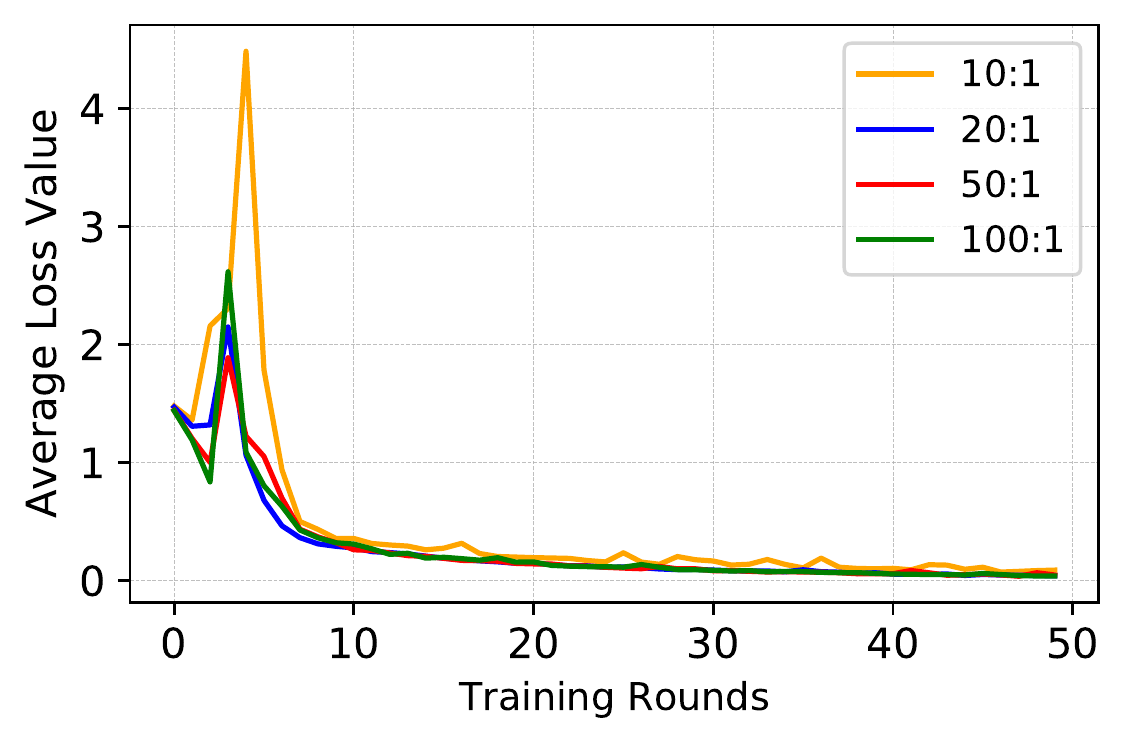}
    \caption{Average loss value of Ratio Loss with different global imbalance ratios during FL training over FEMNIST.}
    \label{Sup_curve_femnist_ratio}
\end{figure}

\begin{figure}[htb]
    \centering
    \includegraphics[width=0.9\linewidth]{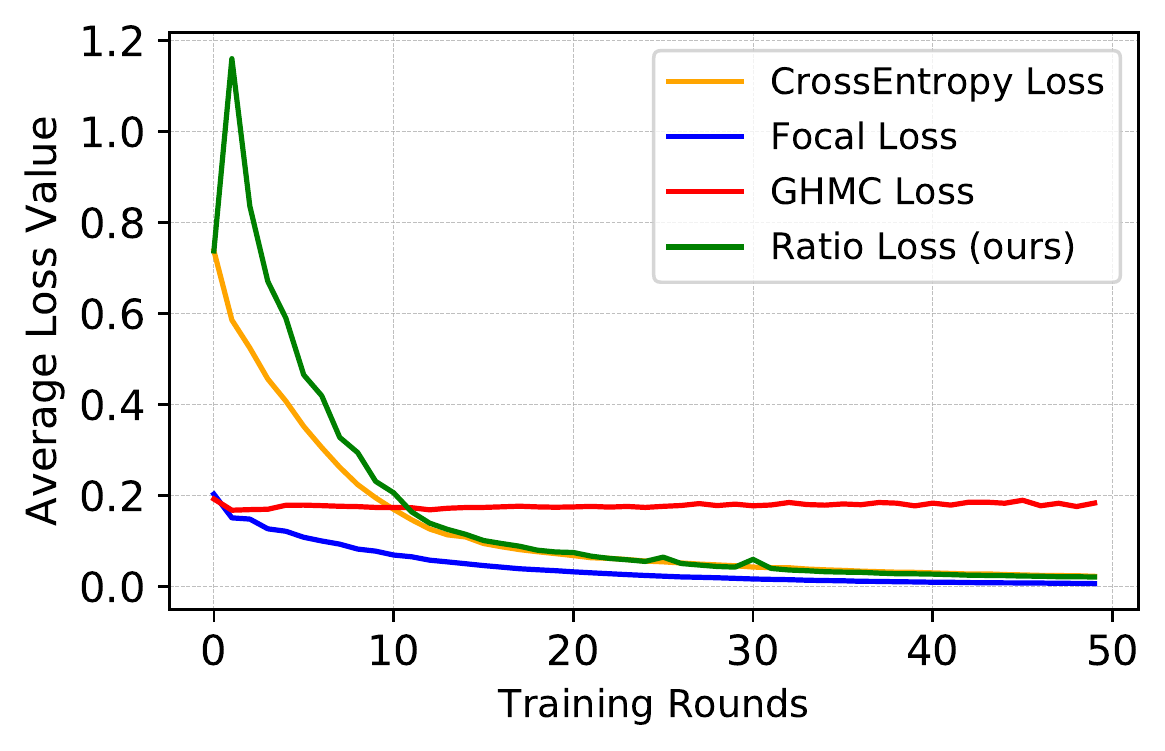}
    \caption{Average loss value of four different loss functions (CrossEntropy Loss, Focal Loss, GHMC Loss and Ratio Loss) with global imbalance ratio $\Gamma = 100:1$ during FL training over CIFAR10.}
    \label{Sup_curve_cifar_four_losses}
\end{figure}

\begin{figure}[htb]
    \centering
    \includegraphics[width=0.9\linewidth]{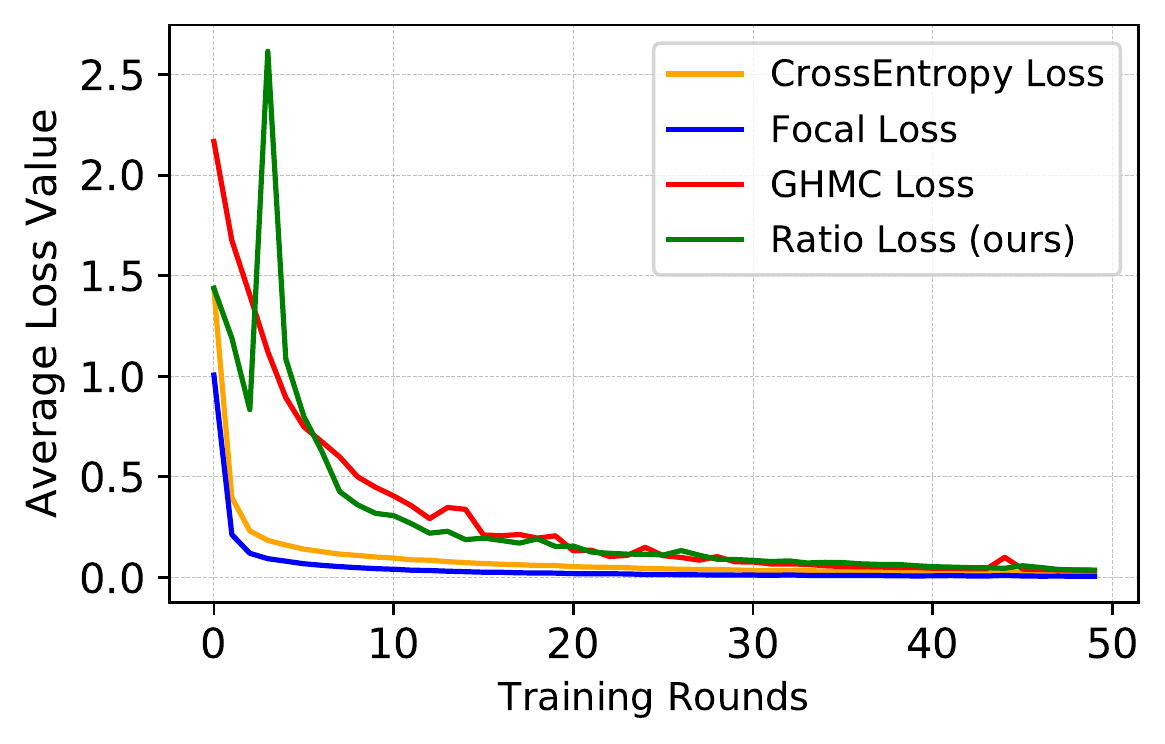}
    \caption{Average loss value of four different loss functions (CrossEntropy Loss, Focal Loss, GHMC Loss and Ratio Loss) with global imbalance ratio $\Gamma = 100:1$ during FL training over FEMNIST.}
    \label{Sup_curve_femnist_four_losses}
\end{figure}

We also include the convergence curves of neural network with different loss functions during FL training. Figs.~\ref{Sup_curve_cifar_ratio} and~\ref{Sup_curve_femnist_ratio} show the changes of training loss for our Ratio Loss with different global class imbalance ratio ($\Gamma = 10:1, 20:1, 50:1, 100:1$) over CIFAR10 and FEMNIST, respectively.  Figs.~\ref{Sup_curve_cifar_four_losses} and~\ref{Sup_curve_femnist_four_losses} show the process of loss degradation for four different losses (Ratio Loss, CrossEntropy Loss, Focal Loss, and GHMC Loss) with the imbalance ratio $\Gamma = 100:1$ over CIFAR10 and FEMNIST, respectively. We can conclude that our Ratio Loss degrades smoothly during imbalanced training, similar to other loss functions.

\smallskip
\noindent \textbf{Dynamic Global Imbalance Ratio}

We also consider the situations where the global imbalance ratio is changing continuously, and conduct experiments to analyze the performance of different loss functions. We do not fix the chosen clients but randomly select them. In this case, the global imbalance ratio is determined by both the prior setting and the specific selection of clients, that is to say, the imbalance ratio is dynamically changing with different chosen clients at each training round. We follow the same setting as previous, and conduct experiments for different loss functions (Ratio Loss, CrossEntropy Loss, Focal Loss, GHMC Loss) with $\Gamma = 10:1, 20:1, 50:1, 100:1$, respectively, over FEMNIST. The detailed results with respect to the AUC score and Ac.M are shown in Table~\ref{Sup_fe_dynamic}. 

\begin{table}[htb]
\centering
\small
\setlength{\tabcolsep}{1.9mm}{
\begin{tabular}{c|c|ccccc}
\hline
\multicolumn{2}{c|}{\textbf{Data}}                                              & \multicolumn{5}{c}{\textbf{FEMNIST}}                                               \\ \hline
\multicolumn{2}{c|}{$\Gamma$}                                                & B.             & 10:1           & 20:1           & 50:1           & 100:1          \\ \hline
\multirow{4}{*}{\textbf{\begin{tabular}[c]{@{}c@{}}Ac.M\\ \%\end{tabular}}} & $L_{C\!E}$ & 84.42          & 36.77          & 06.73          & 00.00          & 00.00          \\
                                                                            & $L_{F\!L}$ & 87.73          & 39.91          & 17.11          & 00.00          & 00.00          \\
                                                                            & $L_{G\!H\!M\!C}$ & 89.27          & 44.46          & 38.80          & 05.16          & 00.09          \\
                                                                            & $L_{R\!L}$(ours) & \textbf{91.22} & \textbf{55.27} & \textbf{48.54} & \textbf{07.70} & \textbf{00.13} \\ \hline
\multirow{4}{*}{\textbf{AUC}}                                               & $L_{C\!E}$ & .9478          & .8895          & .8477          & .8386          & .8383          \\
                                                                            & $L_{F\!L}$ & .9523          & .8914          & .8604          & .8364          & .8363          \\
                                                                            & $L_{G\!H\!M\!C}$ & .9601          & .9006          & .8928          & .8449          & .8392          \\
                                                                            & $L_{R\!L}(ours)$ & \textbf{.9687} & \textbf{.9181} & \textbf{.9080} & \textbf{.8524} & \textbf{.8413} \\ \hline
\end{tabular}
}
\caption{Comparison between our method ($L_{R\!L}$) and previous methods based on CrossEntropy Loss ($L_{C\!E}$), Focal Loss ($L_{F\!L}$) and GHMC Loss ($L_{G\!H\!M\!C}$) in federate learning, over FEMNIST when the global class imbalance ratio is dynamically changing with different selection of clients at each training round.}
\label{Sup_fe_dynamic}
\end{table}

\smallskip
\noindent \textbf{Impact of Mismatch (Additional Results)}

In our main paper, we demonstrate the impact of the mismatch between local imbalance and global imbalance. Here we report more experimental results on this aspect, as shown in Table~\ref{Sup_data_MNIST} for MNIST, Table~\ref{Sup_data_CIFAR10} for CIFAR10, and Table~\ref{Sup_data_Fer2013} for Fer2013. Across all of these experiments, our approach with the Ratio Loss function $L_{R\!L}$ performs the best, with respect to the improvement on AUC score and accuracy on minority classes (Ac.M).

\begin{table*}[htb]
\centering
\small
\setlength{\tabcolsep}{1.45mm}{
\begin{tabular}{c|c|ccccc|ccccc|ccccc}
\hline
\multicolumn{2}{c|}{\textbf{MNIST}}                                              & \multicolumn{5}{c|}{\textbf{$C\!S$=0.6960}}                                            & \multicolumn{5}{c|}{\textbf{$C\!S$=0.6158}}                                            & \multicolumn{5}{c}{\textbf{$C\!S$=0.3984}}                                             \\ \hline
\multicolumn{2}{c|}{$\Gamma$}                                             & B.             & 10:1           & 20:1           & 50:1           & 100:1          & B.             & 10:1           & 20:1           & 50:1           & 100:1          & B.             & 10:1           & 20:1           & 50:1           & 100:1          \\ \hline
\multirow{4}{*}{\textbf{\begin{tabular}[c]{@{}c@{}}Ac.M\\ \%\end{tabular}}} & $L_{C\!E}$ & 98.33          & 90.97          & 86.59          & 75.43          & 64.62          & \textbf{98.24} & 93.09          & 86.87          & 78.57          & 54.82          & 97.98          & 82.78          & 73.41          & 50.12          & 28.93          \\
                                                                            & $L_{F\!L}$ & \textbf{98.48} & 92.41          & 86.44          & 73.46          & 63.91          & 98.13          & 90.55          & 84.44          & 68.82          & 56.18          & \textbf{97.99} & 71.89          & 58.80          & 49.16          & 26.07          \\
                                                                            & $L_{G\!H\!M\!C}$ & 98.01          & 84.17          & 68.20          & 46.84          & 15.07          & 97.77          & 81.60          & 62.68          & 53.06          & 12.95          & 96.78          & -              & -              & -              & -              \\
                                                                            & $L_{R\!L}$(ours) & 98.37          & \textbf{93.02} & \textbf{87.41} & \textbf{76.79} & \textbf{65.18} & 98.22          & \textbf{93.61} & \textbf{88.13} & \textbf{81.27} & \textbf{64.58} & 97.93          & \textbf{85.54} & \textbf{81.41} & \textbf{57.95} & \textbf{38.48} \\ \hline
\multirow{4}{*}{\textbf{AUC}}                                               & $L_{C\!E}$ & .9917          & .9817          & .9724          & .9569          & .9356          & .9901          & .9842          & .9692          & .9606          & .9202          & .9883          & .9588          & .9434          & .9034          & .8630          \\
                                                                            & $L_{F\!L}$ & \textbf{.9924} & .9821          & .9725          & .9517          & .9346          & .9900          & .9765          & .9660          & .9412          & .9202          & \textbf{.9892} & .9339          & .9119          & .8918          & .8534          \\
                                                                            & $L_{G\!H\!M\!C}$ & .9898          & .9649          & .9393          & .9024          & .8484          & .9885          & .9573          & .9237          & .9067          & .8386          & .9774          & -              & -              & -              & -              \\
                                                                            & $L_{R\!L}$(ours) & .9916          & \textbf{.9859} & \textbf{.9763} & \textbf{.9591} & \textbf{.9393} & \textbf{.9903} & \textbf{.9850} & \textbf{.9763} & \textbf{.9657} & \textbf{.9366} & .9890          & \textbf{.9678} & \textbf{.9518} & \textbf{.9101} & \textbf{.8899} \\ \hline
\end{tabular}
}
\caption{Comparison between our method ($L_{R\!L}$) and previous methods based on CrossEntropy Loss ($L_{C\!E}$), Focal Loss ($L_{F\!L}$) and GHMC Loss ($L_{G\!H\!M\!C}$) in federate learning, over three datasets and different levels of global imbalance.}
\label{Sup_data_MNIST}
\end{table*}

\begin{table*}[htb]
\centering
\small
\setlength{\tabcolsep}{1.45mm}{
\begin{tabular}{c|c|ccccc|ccccc|ccccc}
\hline
\multicolumn{2}{c|}{\textbf{CIFAR10}}                                           & \multicolumn{5}{c|}{\textbf{$C\!S$=0.6960}}                                            & \multicolumn{5}{c|}{\textbf{$C\!S$=0.6158}}                                            & \multicolumn{5}{c}{\textbf{$C\!S$=0.3984}}                                             \\ \hline
\multicolumn{2}{c|}{$\Gamma$}                                                 & B.             & 10:1           & 20:1           & 50:1           & 100:1          & B.             & 10:1           & 20:1           & 50:1           & 100:1          & B.             & 10:1           & 20:1           & 50:1           & 100:1          \\ \hline
\multirow{4}{*}{\textbf{\begin{tabular}[c]{@{}c@{}}Ac.M\\ \%\end{tabular}}} & $L_{C\!E}$ & 70.22          & 43.40          & 28.50          & 13.10          & 05.40          & 61.04          & 34.27          & 23.87          & 08.67          & 03.10          & 49.86          & 18.77          & 01.70          & 00.40          & 00.13          \\
                                                                            & $L_{F\!L}$ & 68.19          & 34.77          & 24.93          & 11.40          & 05.10          & 59.45          & 29.60          & 19.43          & 09.07          & 03.47          & 47.35          & 21.03          & 12.13          & 02.37          & 00.80          \\
                                                                            & $L_{G\!H\!M\!C}$ & 68.94          & 39.27          & 26.13          & 09.53          & 04.90          & 57.88          & 34.73          & 20.63          & 09.50          & \textbf{03.97} & 49.37          & 22.00          & 11.83          & \textbf{03.40} & 01.80          \\
                                                                            & $L_{R\!L}$(ours) & \textbf{73.01} & \textbf{47.30} & \textbf{31.57} & \textbf{15.77} & \textbf{05.74} & \textbf{61.97} & \textbf{39.33} & \textbf{25.03} & \textbf{10.68} & 03.90          & \textbf{54.39} & \textbf{22.20} & \textbf{15.10} & 03.17          & \textbf{02.40} \\ \hline
\multirow{4}{*}{\textbf{AUC}}                                               & $L_{C\!E}$ & .8230          & .7793          & .7553          & .7338          & .7215          & .7564          & .7421          & .7257          & .7069          & .6957          & .7023          & .6106          & .6025          & .6185          & .6147          \\
                                                                            & $L_{F\!L}$ & .8128          & .7368          & .7217          & .7046          & .6927          & .7497          & .7086          & .6923          & .6782          & .6688          & .6978          & .6429          & \textbf{.6353} & \textbf{.6275} & \textbf{.6179} \\
                                                                            & $L_{G\!H\!M\!C}$ & .8170          & .7641          & .7512          & .7217          & .7164          & .7475          & .7354          & .7189          & .6991          & .6926          & .6993          & \textbf{.6529} & .6348          & .6249          & .6116          \\
                                                                            & $L_{R\!L}$(ours) & \textbf{.8393} & \textbf{.7985} & \textbf{.7735} & \textbf{.7515} & \textbf{.7345} & \textbf{.7581} & \textbf{.7661} & \textbf{.7438} & \textbf{.7243} & \textbf{.7062} & \textbf{.7211} & .6221          & .6268          & .6193          & .6110          \\ \hline
\end{tabular}
}
\caption{Comparison between our method ($L_{R\!L}$) and previous methods based on CrossEntropy Loss ($L_{C\!E}$), Focal Loss ($L_{F\!L}$) and GHMC Loss ($L_{G\!H\!M\!C}$) in federate learning, over three datasets and different levels of global imbalance.}
\label{Sup_data_CIFAR10}
\end{table*}

\begin{table*}[h!]
\centering
\small
\setlength{\tabcolsep}{1.45mm}{
\begin{tabular}{c|c|ccccc|ccccc|ccccc}
\hline
\multicolumn{2}{c|}{\textbf{Fer2013}}                                           & \multicolumn{5}{c|}{\textbf{$C\!S$=0.9343}}                                            & \multicolumn{5}{c|}{\textbf{$C\!S$=0.8489}}                                            & \multicolumn{5}{c}{\textbf{$C\!S$=0.7411}}                                             \\ \hline
\multicolumn{2}{c|}{$\Gamma$}                                                 & B.             & 10:1           & 20:1           & 50:1           & 100:1          & B.             & 10:1           & 20:1           & 50:1           & 100:1          & B.             & 10:1           & 20:1           & 50:1           & 100:1          \\ \hline
\multirow{4}{*}{\textbf{\begin{tabular}[c]{@{}c@{}}Ac.M\\ \%\end{tabular}}} & $L_{C\!E}$ & \textbf{99.02} & 63.62          & 42.18          & 19.13          & 10.57          & \textbf{98.14} & 55.72          & 35.54          & \textbf{16.05} & 08.02          & \textbf{98.03} & 42.68          & 25.67          & 11.01          & 04.76          \\
                                                                            & $L_{F\!L}$ & 96.23          & 61.74          & 40.38          & 18.99          & 08.95          & 94.47          & 53.95          & 31.63          & 14.05          & 06.96          & 92.35          & 41.63          & 24.68          & 09.16          & 04.36          \\
                                                                            & $L_{G\!H\!M\!C}$ & 95.40          & 63.26          & 42.78          & 17.83          & 08.79          & 84.24          & 56.14          & 35.98          & 14.42          & 06.80          & 67.73          & 42.75          & 25.73          & 09.52          & 04.20          \\
                                                                            & $L_{R\!L}$(ours) & 98.98          & \textbf{63.96} & \textbf{42.93} & \textbf{20.00} & \textbf{10.81} & 98.11          & \textbf{56.20} & \textbf{36.39} & 15.59          & \textbf{08.08} & 98.00          & \textbf{43.23} & \textbf{25.80} & \textbf{11.86} & \textbf{05.21} \\ \hline
\multirow{4}{*}{\textbf{AUC}}                                               & $L_{C\!E}$ & \textbf{.9900} & .8983          & .8414          & .7848          & .7620          & \textbf{.9898} & .8756          & .8229          & .7746          & .7556          & \textbf{.9897} & .8413          & .7977          & .7619          & .7467          \\
                                                                            & $L_{F\!L}$ & .9809          & .8929          & .8368          & .7848          & .7576          & .9703          & .8721          & .8141          & .7698          & .7526          & .9671          & .8390          & .7961          & .7575          & .7456          \\
                                                                            & $L_{G\!H\!M\!C}$ & .9793          & .8943          & .8416          & .7793          & .7570          & .9475          & .8757          & .8223          & .7706          & .7524          & .8932          & .8415          & \textbf{.7989} & .7591          & .7451          \\
                                                                            & $L_{R\!L}$(ours) & .9899          & \textbf{.8989} & \textbf{.8428} & \textbf{.7856} & \textbf{.7628} & .9896          & \textbf{.8770} & \textbf{.8252} & \textbf{.7759} & \textbf{.7622} & .9891          & \textbf{.8423} & .7988          & \textbf{.7626} & \textbf{.7475} \\ \hline
\end{tabular}
}
\caption{Comparison between our method ($L_{R\!L}$) and previous methods based on CrossEntropy Loss ($L_{C\!E}$), Focal Loss ($L_{F\!L}$) and GHMC Loss ($L_{G\!H\!M\!C}$) in federate learning, over three datasets and different levels of global imbalance.}
\label{Sup_data_Fer2013}
\end{table*}

\subsection{Machine Learning Reproduction Details}
All experiments are conducted on a server running Ubuntu 18.04 LTS, equipped with a 2.10GHz CPU Intel Xeon(R) Gold 6130, 64GB RAM, and NVIDIA TITAN RTX GPU cards. 

\smallskip
\noindent \textbf{Data Splitting}
\begin{figure*}[h!]
    \centering
    \includegraphics[width=1.\linewidth]{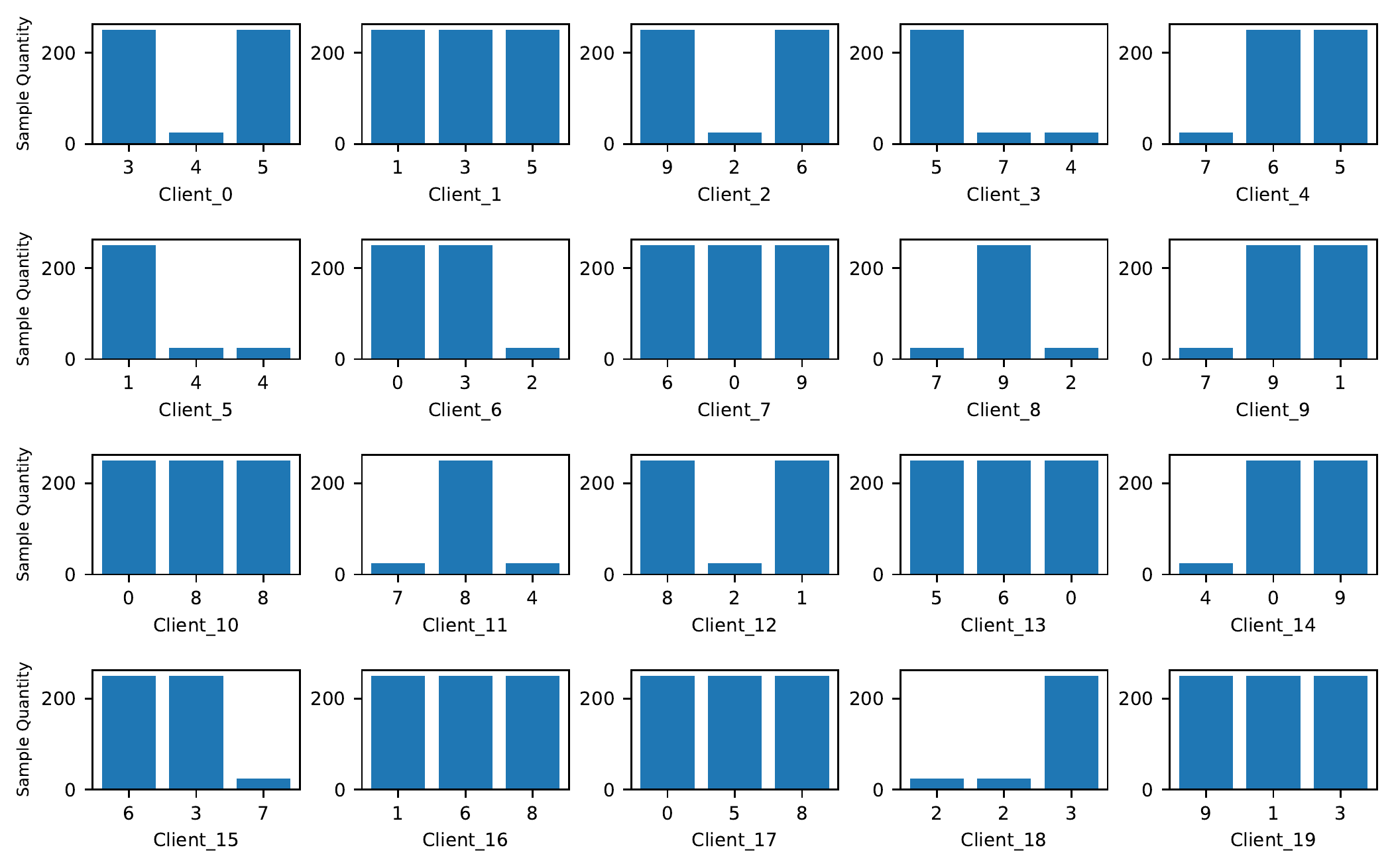}
    \caption{The data compositions of 20 chosen clients when the global imbalance ratio is set as $\Gamma=10:1$ and minor classes is digital 2, 4 and 7, over MNIST.}
    \label{Sup_data_splitting}
\end{figure*}
For datasets of MNIST, CIFAR10 and Fer2013, our data splitting strategy is similar. Here we take MNIST as the representative example. We split 80\% of samples as the training data and 20\% as the testing set. Thus the training set includes approximately 50,000 data samples, and each class corresponds to 5,000 samples. As introduced in the Evaluation section of the main paper, we allocate $1 \sim Q$ (randomly select) classes to 100 participants, and the specific classes are also randomly determined. For each class of a particular participant, we allocate 50 data samples and such allocation is without replacement. The other two datasets are allocated to participants as the strategy as that of MNIST. For the experiments of comparing different loss functions under various global imbalance ratios, we fix the ratio via selecting the same 20 clients and limit each client to own $3 \sim 6$ classes corresponding to different mismatch levels. To build more extreme imbalanced scenarios, we allocate 250 samples to each majority class, and 25 samples to every minority class when the prior ratio setting is $10:1$. Minority classes are randomly determined as $\text{digital}\, 2, 4, 7$. Please refer to Fig.~\ref{Sup_data_splitting} for the visualization of the splitting strategy. 

For FEMNIST, the benchmark has separated it into the training set and the testing set. As the data samples of each writer is relatively few, we group 20 writers into a new client, and thus we change 2,000 writers into 100 clients. For the comparison experiments of loss functions, we set the former 7 classes as the minority classes in advance. If the original sample number of a minority class $\mathcal{B}$ is $N_o$, we will randomly select part of chosen clients and remove their data samples of class $\mathcal{B}$, finally setting the quantity of remaining samples as $N_o / \Gamma$.

\smallskip
\noindent \textbf{Auxiliary Data}

The auxiliary data we use plays a role like a set of validation data, and it is consisted by a small number of  data samples ($32$ in our experiments) for different classes. Theoretically, we only need one data sample for each class to utilize it for computing the gradient. For the current design of our monitoring scheme, we assume that there is no significant distribution discrepancy between the auxiliary data and the training data owned by clients. Note that in the cases where there is significant distribution gap between the training data and the auxiliary one, we can try to incorporate domain adaptation to our monitoring scheme. 

In our experiments, we compose the auxiliary data from the testing dataset, and the selection of samples is random for MNIST, CIFAR10 and Fer2013. As for FEMNIST, in order to maintain the feature heterogeneity, we compose its auxiliary data from the other writers (2000 $\sim$ 3600) rather than the participated writers (0 $\sim$ 2000). Therefore, in practical scenarios, we do not need some clients to share the server with their data to compose the auxiliary data. Instead, we can collect it from the public data, or apply some reproduction and augmentation technologies, e.g., GAN~\cite{ledig2017photo}, to synthesize the auxiliary data based on the prior knowledge of FL administrator or the privileged features of classes~\cite{vapnik2015learning}.

\subsection{Application Scenarios}

In order to comprehend the practical scenarios of our monitoring scheme, we consider two problems -- standard Federated Learning (FL) and Federated Transfer Learning (FTL). Standard FL is often built for the situations where the features of training data across participated clients are similar, i.e., the difference between features of different clients is not very significant. And for FL, the local models are the same as the global one. 
However, in the cases where the difference between features of different clients becomes significant, i.e., when there is obvious distribution discrepancy among data sets owned by different clients, FL may incorporate the techniques of transfer learning and become FTL~\cite{liu2018secure}. And the structures of local models are usually different and specific to each domain in this case~\cite{peterson2019private, peng2019federated}. 

In this paper, we focus on the class imbalance problem for standard FL, where the difference between features of different clients is not very significant. This is common in many practical applications, such as key-board typing~\cite{hard2018federated, ramaswamy2019federated}, commands for autonomous driving~\cite{samarakoon2018federated}, and abnormal heart rate monitoring by wearable devices~\cite{nguyen2019diot}. Note that empirically, for some cases where there is significant feature difference across clients (e.g,. FEMNIST is made up of digital signatures for approximately 3,600 different writers and thus there are various writing styles), our monitoring scheme still performs very well, as shown in the experimental results reported earlier. For more thorough and general study of the cases where the feature difference across clients is significant, we plan to address them in the future work as the class imbalance problem for FTL. 

As for Ratio Loss, it not only performs the best in the setting of FL, but also outperforms other state-of-the-art methods in regular training without FL, as shown in the experiments introduced in the main paper. And as stated there, using sample quantities to define the imbalance problem is less effective than using the contribution to gradients of different samples, especially for the cases with relatively significant feature heterogeneity. In addition, different selections of clients at each round may have different and highly-biased data compositions. If the future composition is consistent with the existing imbalance, our Ratio Loss function will try to invert this imbalance trend as much as possible, i.e., by enlarging the contribution of gradients from minority classes while shrinking that from majority classes. If the future data composition is inconsistent with the current imbalance, our Ratio Loss function will try to leverage it for mitigating the imbalance. As shown in the experiments earlier in the supplementary material and mentioned in the main paper, our Ratio Loss function performs very well when the imbalance is dynamically changing.